\pdfoutput=1  
\documentclass[twoside,11pt]{article}

\usepackage[preprint]{jmlr2e}

\usepackage{amsmath}
\usepackage{booktabs}
\usepackage{array}
\usepackage{algorithm}
\usepackage{algpseudocode}

\usepackage{hyphenat}
\usepackage{microtype}
\newcommand{\err}{\mathrm{err}}
\newcommand{\ind}{\mathbf{1}}

\usepackage{lastpage}
\jmlrheading{23}{2026}{1-\pageref{LastPage}}{1/21; Revised 5/22}{9/22}{21-0000}{P. Brandimarte, N. Aranjuelo, M. Nieto and O. Otaegui}

\ShortHeadings{Adaptive Conformal Filtering for Hardware-Aware NAS}{Brandimarte, Aranjuelo, Nieto and Otaegui}
\firstpageno{1}

\begin{document}

\title{Adaptive Conformal Filtering for Hardware-Aware\\
Neural Architecture Search: Coverage and Pruning Trade-offs}

\author{\name Pedro Brandimarte$^{1, \ast}$ \email pbrandimarte@vicomtech.org
       \AND
       \name Nerea Aranjuelo$^1$ \email naranjuelo@vicomtech.org
       \AND
       \name Marcos Nieto$^1$ \email mnieto@vicomtech.org
       \AND
       \name Oihana Otaegui$^1$ \email ootaegui@vicomtech.org \\
       \addr $^1$Vicomtech Foundation, Basque Research and Technology Alliance (BRTA)\\
       Donostia-San Sebasti\'an, Spain\\[4pt]
       \addr $^{\ast}$Corresponding author}

\editor{Editor Name}

\maketitle

\begin{abstract}
Conformal filtering can reduce the cost of hardware-aware neural architecture search by discarding candidates before evaluation, but changing proposal distributions challenge static calibration.
We analyze what adaptive coverage control does, and does not, provide in reinforcement-learning-based search.
Building on Adaptive Conformal Inference (ACI), we compare global, step-size-aggregated, candidate-dependent, and depth-conditional calibration to distinguish coverage tracking from pruning efficiency and search quality.
Across three CNN families, ACI tracks requested update-time coverage on evaluated proxy outcomes with a mean absolute calibration error of approximately $10^{-3}$, consistent with its deterministic guarantee, and prunes 25--50\% of evaluations without observed loss in partial-training proxy accuracy at the selected thresholds.
Our central finding is that even tightly calibrated coverage does not prevent substantial false discards under a realistic low-fidelity proxy: on NAS-Bench-201 with measured hardware costs, reusing the full-fidelity threshold with a 12-epoch proxy produces high false-discard rates.
A retrospective threshold sweep reduces these rates at the cost of less pruning, with broadly similar trade-offs across filters.
A conservative Gaussian-process baseline remains competitive despite overcoverage.
Sequential audits show depth-specific adaptation reduces coverage imbalances where groups are sufficiently sampled.
An exploratory acquisition study attributes most of its gain to surrogate-based selection rather than online calibration.
The study delineates where adaptive calibration helps and where threshold placement and proxy fidelity remain decisive.
Source code: \url{https://github.com/Vicomtech/rl-hw-nas}.
\end{abstract}


\section{Introduction}
\label{sec:intro}
Deploying deep neural networks on edge devices requires meeting tight memory and latency budgets, from microcontrollers with a few hundred kilobytes of SRAM to embedded accelerators with real-time constraints.
On-device inference reduces communication latency, cost, energy use, and privacy exposure.
Designing for each device nevertheless requires balancing accuracy, memory footprint, inference latency, and quantization precision.
Depth, width, kernel sizes, connectivity, and per-layer bit-widths produce millions of configurations, whose accuracy must be estimated through training or a proxy.
Hardware-aware neural architecture search (HW-NAS) automates this exploration by optimizing a scalar reward that combines task accuracy with hardware penalties, and has repeatedly produced edge models that dominate hand-designed baselines \citep{Tan2019, Wu2019, Cai2020, Lin2020}.

Search strategies include reinforcement learning (RL) \citep{Zoph2017, Tan2019}, evolution \citep{Real2019a}, differentiable optimization \citep{Liu2018}, weight-sharing supernets \citep{Cai2020}, and Bayesian optimization (BO) \citep{White2021}.
Whatever the proposal mechanism, a common bottleneck dominates the wall-clock cost: \textbf{candidate evaluation}.
In searches that train candidates individually, a realistic run consumes hundreds to thousands of partial-training evaluations.
Many evaluations are spent on uncompetitive candidates, motivating reward predictors, early-stopping proxies, and weight sharing alongside better search strategies.
Well-tuned random search also remains a strong baseline on moderate spaces \citep{Li2020d}, making efficient allocation of the evaluation budget a central concern.

MARCO \citep{Fayyazi2026} addresses this cost with a conformal filter inside a multi-agent RL search over resource-constrained edge CNN spaces.
A surrogate predicts each candidate's reward, and a split-conformal offset \citep{Vovk2022, Papadopoulos2002} converts the prediction into an upper bound.
Under exchangeability, this bound covers the true reward with probability at least $1-\delta$; candidates whose bounds fall below an acceptance threshold are discarded.
The coverage guarantee is distribution-free: it requires no assumption about the surrogate's error distribution.
It does, however, require \textbf{exchangeability} between calibration and test candidates.

That assumption is violated by the search itself.
As the RL policy learns, its proposal distribution departs from the warm-up distribution used for calibration.
Sequential construction adds distribution shift within each episode as partial networks grow layer by layer.
Under this drift, realized coverage can depart from the target $1-\delta$, with the direction and size of the deviation depending on the calibration data.
Over-coverage can waste evaluations by retaining uncompetitive candidates; under-coverage can falsely discard strong candidates, potentially including the best architecture.
Periodic recalibration mitigates distribution shift.
In our experiments, this reduces variability but does not restore coverage control.
A global offset also leaves depth-dependent miscoverage unresolved.

Using a common experimental framework, we examine how online coverage control relates to pruning efficiency and search quality across uncertainty models and proposal mechanisms.
Adaptive Conformal Inference (ACI, \cite{Gibbs2021}) does not estimate a quantile once but \textit{controls} the coverage error by online feedback, updating the conformal offset after every evaluation.
Its guarantee is of a different kind: long-run coverage converges to $1-\delta$ for \textit{arbitrary bounded-residual} candidate sequences under an uninterrupted recursion, without exchangeability or stationarity.
We aggregate experts with different step sizes \citep{Zaffran2022}, use normalized scores to obtain candidate-dependent widths, and apply a Mondrian construction for depth-conditional coverage.
The resulting optimistic bound can also guide exploration.
Greedy acquisition selects candidates with the largest bounds, following the optimism principle of the Gaussian-process upper confidence bound (GP-UCB) \citep{Srinivas2010} with coverage controlled online.
An AlphaZero-style planner \citep{Silver2018} instead uses the bound to value simulated completions of partial networks.

Our contributions are as follows:

\begin{enumerate}
    \item \textbf{A comparative analysis of adaptive conformal filtering under search-induced drift.}
    We bring ACI, step-size aggregation, and candidate-dependent and depth-conditional uncertainty models into single-step and sequential hardware-aware search under common reward and hardware interfaces, and examine how coverage tracking relates to pruning efficiency and search quality.
    \item \textbf{Coverage control across candidates and construction depths.}
    Across three CNN families, we characterize update-time coverage on evaluated candidates and audit depth-dependent imbalances hidden by marginal coverage.
    Per-depth adaptation reduces conditional error substantially on LeNet-like and MobileNet-like networks; sparse ResNet-like groups reveal its finite-sample limits.
    \item \textbf{Quantitative analysis of pruning and search decisions.}
    On NAS-Bench-201 with measured hardware costs, a retrospective threshold sweep shows that reusing the full-fidelity threshold with a 12-epoch proxy produces high false-discard rates, and that lower proxy-scale thresholds reduce them at the cost of less pruning, with broadly similar trade-offs across filters.
    Budget-matched controller comparisons and exploratory acquisition ablations on one constrained testbed distinguish the observed gains from surrogate-based selection, optimism, and online calibration.
\end{enumerate}

\section{Related Work}
\label{sec:sota}
\subsection{Hardware-aware neural architecture search}

Hardware-aware NAS has grown into a broad subfield \citep{Baymurzina2022}, surveyed and taxonomized by search space, search strategy and hardware cost-estimation approach \citep{Benmeziane2021}.
Architecture search began as an accuracy-only problem solved with RL at high computational cost \citep{Zoph2017, Baker2017}, before incorporating deployment constraints.
MnasNet \citep{Tan2019} folded measured mobile latency into the reward, and ProxylessNAS \citep{Cai2019} and FBNet \citep{Wu2019} made latency differentiable through per-operator lookup tables.
Once-for-All \citep{Cai2020} decoupled training from search by extracting sub-networks from a single weight-sharing supernet, and MCUNet \citep{Lin2020} pushed the setting to microcontrollers, co-designing the architecture with the inference engine under kilobyte-scale memory budgets.
Differentiable search over relaxed spaces \citep{Liu2018} offers an alternative, though its stability and coupling to a specific supernet limit portability across cost models.
Evolutionary NAS treats resource budgets as explicit constraints \citep{Liu2023}.
Recent frameworks adapt constraint handling during search: Xie et al.~\cite{Xie2026}, for example, increase a Params/FLOPs penalty across generations and use structured filter pruning as a local operator.
Our work operates in resource-constrained CNN spaces with per-layer quantization choices, evaluated against SRAM and latency budgets, but the contribution is orthogonal to search-space or cost-model design: we target the statistical machinery that decides which candidates are worth evaluating, and it applies unchanged across the families and oracles we test.

\subsection{Reducing the evaluation cost}

Since candidate training dominates NAS wall-clock time, a large literature attacks evaluation cost directly.
Partial-training proxies truncate epochs and data; weight sharing amortizes training across candidates \citep{Pham2018, Cai2020}; learned performance predictors rank candidates from architecture features \citep{White2021, Wen2020a}; zero-cost proxies score untrained networks from single minibatches \citep{Mellor2021, Abdelfattah2021}.
Multi-fidelity methods such as Hyperband \citep{Li2017b} allocate evaluation resources adaptively, increasing training epochs or data for promising candidates and stopping others early.
That line is orthogonal to ours: it decides \textit{how much} to spend on a candidate, whereas our filter decides, with a calibrated error statement, \textit{whether} to spend at all, and the two compose.
Point predictors alone provide no calibrated coverage statement.
Errors on unseen regions can cause strong candidates to be discarded, a failure that average prediction-error metrics may conceal.
Our filter wraps a small multilayer perceptron (MLP) in a conformal bound with controlled coverage.
The predictor supplies reward estimates; calibration determines the bound, and false-discard performance is evaluated separately.
A distinct, complementary line \textit{precomputes} the reward instead of approximating it: tabular NAS benchmarks tabulate ground-truth accuracy for an entire fixed search space (the topology search space of NATS-Bench, \cite{Dong2021}, also known as NAS-Bench-201), and HW-NAS-Bench \citep{Li2021b} adds \textit{measured} on-device latency and energy for the same architectures.
These tables reveal the rewards of pruned candidates, allowing direct false-discard measurement with measured hardware costs (Sec.~\ref{ssec:e3}).

\subsection{RL controllers and random-search baselines}

RL controllers for NAS span single-agent policy gradient \citep{Zoph2017}, weight sharing \citep{Pham2018}, and Q-learning over layer sequences \citep{Baker2017}.
MARCO \citep{Fayyazi2026} uses cooperating agents under centralized training / decentralized execution (CTDE; cf. MADDPG, \cite{Lowe2017}; MAPPO, \cite{Yu2022a}).
A hardware\hyp{}configuration agent selects architecture parameters, and a quantization agent selects per-layer bit-widths.
Proximal policy optimization (PPO) trains both agents on a hardware-aware reward as they construct candidates layer by layer.
Split-conformal bounds filter candidates before evaluation.
Our filtering study reproduces this controller, including its layer-by-layer formulation and per-step credit assignment.
A parallel evaluation literature, however, has repeatedly found that well-tuned random search matches or beats learned NAS controllers on moderate spaces \citep{Li2020d, Yang2020}.

\subsection{Bayesian optimization and optimism-based search}

Bayesian optimization is the canonical sample-efficient search: a probabilistic surrogate (typically a Gaussian process, GP) plus an acquisition rule, with GP-UCB supplying optimism in the face of uncertainty and regret guarantees under correct model specification \citep{Srinivas2010}.
BO has been adapted to architecture spaces through tailored kernels \citep{Kandasamy2018} and neural predictors with ensemble uncertainty \citep{White2021}.
GP uncertainty depends on the kernel and noise assumptions, so its nominal coverage need not hold under distribution shift.
Closest to our acquisition track, \cite{Stanton2023} combine BO with conformal prediction sets to protect acquisition decisions against \textit{feedback covariate shift}, the train/test distribution shift a design algorithm induces by choosing where to query \citep{Fannjiang2022}.
Our setting differs in the drift source (an improving RL policy and within-episode depth shift, not only the feedback shift induced by acquisition itself), in the calibration mechanism (ACI's feedback control rather than full conformal Bayes with covariate-shift weighting), and in the search space (discrete layer-by-layer construction, where the bound also drives tree search).

\subsection{Tree search for architecture construction}

Monte-Carlo tree search (MCTS) has been applied to NAS by AlphaX \citep{Wang2020b}, which guides MCTS with a learned value predictor, and LaNAS \citep{Wang2021b}, which learns to partition the space to focus sampling.
These works establish that the layer-by-layer construction process is a natural planning tree; they evaluate leaves with surrogate predictors but attach no calibrated statement to the values that guide exploration, and they spend real evaluations by fixed schedules.
Our conformal-optimistic MCTS uses an AlphaZero-style planner \citep{Silver2018}, with controller priors and the PUCT (predictor-weighted upper confidence bounds for trees) selection rule.
It values simulated completions using a calibrated optimistic bound and trains only committed candidates.

\subsection{Conformal prediction on non-exchangeable data}

Conformal prediction converts any predictor into set- or interval-valued predictions with finite-sample marginal coverage, assuming only exchangeability \citep{Vovk2022, Shafer2008}; the split (inductive) construction makes it cheap enough to deploy around learned models \citep{Papadopoulos2002, Lei2018}.
Locally-adaptive scores normalize by a difficulty estimate to obtain per-input interval widths \citep{Papadopoulos2008}, Mondrian (group-conditional) constructions give per-group validity at the cost of per-group data \citep{Vovk2022, Bostroem2020}, and recent work characterizes the interpolation between marginal and fully conditional guarantees \citep{Gibbs2025}.
The frontier relevant to us is coverage \textit{without} exchangeability: reweighting handles known covariate shift \citep{Tibshirani2019}, robust variants bound the coverage gap under bounded drift \citep{Barber2023}, and Adaptive Conformal Inference \citep{Gibbs2021} abandons distributional assumptions altogether, controlling long-run coverage on arbitrary sequences by online updates.
Follow-up work aggregates experts to avoid selecting a single step size \citep{Zaffran2022}, strengthens regret and coverage dynamics \citep{Bhatnagar2023, Gibbs2024, Angelopoulos2023}, and studies decaying step sizes that trade drift tracking for convergence \citep{Angelopoulos2024}.
We retain a constant step size to track persistent drift.
Applications of conformal methods inside sequential decision loops are emerging: safe autonomous decisions \citep{Lekeufack2024}, online prediction under semi-bandit feedback \citep{Ge2025}, conformal-prediction-driven active learning \citep{Matiz2020}, and conformal BO \citep{Stanton2023}.
The semi-bandit setting shares a censored-feedback structure with our own filter, which observes rewards only for the candidates it accepts.
Conformal methods also guide black-box and hyperparameter optimization: \cite{Salinas2023} combine split conformalized quantile regression with Thompson sampling, with evaluations on NAS benchmarks.
Our study applies ACI feedback updates to conformal filtering in hardware-aware NAS, controlling long-run coverage on the evaluated stream under policy drift, and investigates depth-conditional calibration during sequential construction.

\section{Methodology}
\label{sec:methods}
We couple a multi-agent RL controller with an adaptive conformal filter, then use the calibrated bound for acquisition.
We describe the search problem, controllers, filters, acquisition mechanisms, evaluation protocol, and coverage guarantees.

\subsection{Problem formulation}
\label{ssec:problem}

\subsubsection{Hardware-aware objective}

Given a model family's discrete design space $\mathcal{A}$ and a target device with a latency budget $T_B$ and a memory budget $M_B$, we seek $a^\star = \arg\max_{a \in \mathcal{A}} R(a)$, where the hardware-aware reward of a candidate architecture `$a$' combines task accuracy with soft hardware penalties:
\begin{equation}
\label{eq:reward}
R(a) = A(a) - \lambda \,\frac{T(a)}{T_B} - \nu \max\!\Big(0, \frac{M(a) - M_B}{M_B}\Big),
\end{equation}
with $A(a) \in [0,1]$ the accuracy under the evaluation proxy (Sec.~\ref{sssec:proxy}), $T(a)$ the estimated inference latency, and $M(a)$ the estimated memory footprint (weights plus peak activations).
Equation~\eqref{eq:reward} follows MARCO's reward with one modification: the memory term is a \textit{continuous} soft barrier rather than a binary indicator, so that the penalty is informative in the neighbourhood of the budget instead of switching discontinuously across it.
The latency penalty applies to every candidate, with $T_B$ providing its reference scale; the memory penalty applies only above $M_B$.
Neither term imposes a hard feasibility constraint.
We use $\lambda = 0.2$ and $\nu = 1.0$; budgets are set per device regime (e.g.\ $T_B = 0.3$\,ms and $M_B = 512$\,kB for the resource-constrained spaces, tightened to $M_B = 256$\,kB on the constrained testbed used for the search-efficiency experiments).

\subsubsection{Search spaces}
\label{sssec:spaces}

We use macro and sequential search formulations across three CNN families, plus a tabular benchmark with exact rewards and measured hardware costs (Sec.~\ref{ssec:e3}).
The macro families are searched in the single-step bandit formulation of Sec.~\ref{sssec:bandit}: they expose the filter to a single drift source (policy improvement) and are cheap enough (one evaluation per candidate) for wide multi-seed, multi-level calibration sweeps.
The per-layer spaces use the sequential Markov decision process (MDP) of Sec.~\ref{sssec:sequential}.
They introduce a second drift source as partial networks grow within each episode, allowing coverage to be assessed conditional on construction depth.

\textbf{Macro (single-decision) families.}
One decision produces a complete architecture.
Three small CNN families span three classic convolutional block types: plain convolutions (a LeNet-like family, MNIST), residual blocks (a ResNet-like family, CIFAR-10), and depthwise-separable convolutions (a MobileNet-like family, CIFAR-100).
The knobs are depth (number of blocks), base channel width, kernel size $\{3,5\}$, skip connections on/off, pooling type $\{\max, \text{avg}\}$, and a per-block quantization bit-width $b_i \in \{4, 8\}$.

\textbf{Per-layer (sequential) families.}
A heterogeneous space in which each layer $t$ carries its own $(\text{kernel}_t, \text{channels}_t, b_t)$, together with global skip/pooling choices; depth ranges are family-specific (LeNet-like 2--4, MobileNet-like 4--12, ResNet-like 8--20).
With a branching factor of roughly 20 choices per layer at depths up to 20, the space contains on the order of $20^{20}$ terminal architectures, and (crucially for calibration) the candidate distribution shifts \textit{within} each episode as the partial network deepens.

\textbf{Benchmark (tabular) space.}
For the tabular benchmark of Sec.~\ref{ssec:e3} we also search the topology search space of NATS-Bench \citep{Dong2021} (the NAS-Bench-201 cell space): a fixed cell of four nodes and six directed edges, each edge taking one of five operations ($15{,}625$ cells), in the single-step formulation (one categorical choice per edge).
NATS-Bench provides full-fidelity (200-epoch) and low-fidelity (12-epoch) CIFAR-100 accuracies; HW-NAS-Bench supplies \textit{measured} edge-device latency and energy \citep{Li2021b}.
The search observes the 12-epoch reward, while false discards are scored against the tabulated 200-epoch reward (Sec.~\ref{ssec:e3}).

\subsubsection{Evaluation proxy and quantization}
\label{sssec:proxy}

We estimate $A(a)$ with a partial-training proxy: each candidate is trained for $E$ epochs on a fixed random fraction $\rho$ of the training set.
The subset is determined by the seed and shared by all candidates.
We then measure top-1 accuracy on the held-out evaluation split.
The default setting is $(E, \rho) = (5, 0.1)$.
A stronger proxy, $(E,\rho)=(15,0.3)$, is used for the MobileNet-like calibration and replay experiments, the ResNet-like controller and acquisition comparisons, and the additional ResNet-like depth audit.
Under the weak proxy, the ResNet-like reward landscape has an isolated optimum typically found during warm-up, making search methods difficult to distinguish.
The stronger proxy provides a landscape on which search can make sustained progress.
Coverage is measured against the proxy outcomes observed in each experiment.
It therefore describes calibration to those outcomes, while solution quality also depends on how faithfully the proxy ranks architectures.

After proxy training, fake post-training quantization (PTQ, \cite{Jacob2018}) quantizes each block's weights to its chosen $b_i$ before scoring.
Lower bit-widths therefore affect both the cost estimate and the measured accuracy.

\subsubsection{Analytical hardware cost model}

Our deterministic MCU-style cost model estimates $(T,M)$ from per-layer parameter and multiply-accumulate counts, scaled by bit-width.
Memory combines weight storage and peak activations, the main constraints on microcontroller deployments \citep{Lin2020}.
For layer $i$ with $c_{in}$ input channels, $c_{out}$ output channels, kernel size $k$, spatial resolution $s_i$ (halved per stage), and bit-width $b_i$, the parameter count is
\begin{equation*}
P_i =
\begin{cases}
c_{in} k^2 + c_{in} c_{out} & \text{depthwise-separable block,}\\[2pt]
c_{in} c_{out} k^2 & \text{plain convolutional block,}\\[2pt]
c_{in} c_{out} k^2 + c_{out}^2 k^2 & \text{residual block,}
\end{cases}
\end{equation*}
(plus $c_{in} c_{out}$ for a residual skip projection), and the memory and latency estimates are
\begin{equation}
\label{eq:cost}
\begin{aligned}
M(a) &= \underbrace{\textstyle\sum_i P_i \cdot \tfrac{b_i}{8}}_{\text{weight bytes}}
 + \underbrace{\textstyle\max_i \; c_{out,i}\, s_i^2 \cdot \tfrac{b_i}{8}}_{\text{peak activations}},\\
T(a) &\propto \textstyle\sum_i P_i\, s_i^2 \cdot \tfrac{b_i}{8}.
\end{aligned}
\end{equation}
The proportionality constant in $T(a)$ is $1.0$\,ms per bit-width-weighted GMAC, placing $T(a)$ on the same millisecond scale as $T_B$.
The model is not calibrated to a specific device, but it is monotone in the ordinal design knobs (width, bit-width, and the budget terms); for categorical choices such as pooling or connectivity we make no monotonicity claim.
This is what the search dynamics and the budget-binding regime require.
A truncated (first-$t$-layers) configuration is costed under the same spatial schedule used to build it, so partial networks in the sequential MDP are costed consistently.
The peak-activation term is a simplification: it takes the single largest layer activation and does not model input and output buffers that are live simultaneously, or residual-branch buffers \citep{Lin2020}.

\subsection{Search controllers}
\label{ssec:controllers}

We use a multi-agent PPO controller and standard search baselines to study how candidate selection affects calibration.
Policy updates generate a changing candidate distribution; sequential construction adds within-episode depth shifts.
The filter is controller-agnostic: it receives only proposed candidates and observed rewards.

\subsubsection{Multi-agent PPO: single-step formulation}
\label{sssec:bandit}

We adapt MARCO's two-agent controller and sequential construction scheme \citep{Fayyazi2026}, evaluating both single-step and layer-by-layer formulations.
Under CTDE, two actors factorize the decision: a \emph{hardware\hyp{}configuration agent} (HCA) samples the macro-architecture knobs, and a \emph{quantization agent} (QA) samples per-block bit-widths autoregressively, conditioned on the HCA's choices.
The single-step formulation is a horizon-1 MDP, or bandit.
One joint decision produces a complete architecture and receives its evaluated reward.
This isolates policy improvement as the source of drift; Sec.~\ref{sssec:sequential} retains the layer-by-layer formulation.
Given a batch of rewards $r_{1:N}$ and a learned scalar baseline $V$, both actors share the normalized advantage
\begin{equation*}
\hat{A}_j = \frac{(r_j - V) - \operatorname{mean}_j(r_j - V)}{\operatorname{std}_j(r_j - V) + \varepsilon},
\end{equation*}
and each actor $m \in \{\text{HCA},\text{QA}\}$ minimizes the PPO clipped-surrogate loss \citep{Schulman2017}:
\begin{equation}
\label{eq:ppo}
\mathcal{L}_m = -\,\mathbb{E}_j\Big[\min\big(\rho_j^{(m)} \hat{A}_j,\;
\operatorname{clip}\big(\rho_j^{(m)}, 1\!-\!\epsilon, 1\!+\!\epsilon\big) \hat{A}_j\big)\Big],
\end{equation}
with $\rho_j^{(m)}=\pi_m(a_j)/\pi_m^{\mathrm{old}}(a_j)$.
The total loss is $\sum_m\mathcal{L}_m+c_v\frac{1}{N}\sum_j(V-r_j)^2-c_H\mathcal{H}[\pi]$, adding value regression and an entropy bonus to discourage premature concentration.
Hyperparameters are listed in Table~\ref{tab:settings}.

\subsubsection{Sequential per-layer MDP}
\label{sssec:sequential}

The sequential formulation builds each network layer by layer.
The state $s_t$ encodes the partial architecture, the layer specifications chosen so far, the current position, and cumulative cost features.
At each step the HCA selects the current layer's (kernel, channels) and the QA its bit-width; the episode terminates at a sampled target depth.
The \textbf{per-step reward} applies Eq.~\eqref{eq:reward} to the partial network built so far, evaluated with the proxy, providing temporal-difference credit assignment.
This optimizes discounted cumulative partial-network reward as a dense surrogate objective, without guaranteeing agreement with terminal-architecture quality.
Final selection uses the reward of the completed architecture.
A state-conditioned critic $V_\phi(s)$ is trained to the one-step temporal-difference (TD) target $r_t + \gamma_{\text{RL}} V_\phi(s_{t+1})$, and the advantage is the corresponding TD(0) error with discount $\gamma_{\text{RL}} = 0.99$:
\begin{equation}
\label{eq:td}
\hat{A}_t = r_t + \gamma_{\text{RL}}\, V_\phi(s_{t+1})\,(1 - d_t) - V_\phi(s_t),
\end{equation}
where $d_t$ indicates episode termination; advantages are normalized as above and each actor minimizes the clipped surrogate of Eq.~\eqref{eq:ppo}.
A \emph{terminal-only} variant evaluates only the completed network, using one real evaluation per episode.
This permits comparison with complete-network search at matched evaluation budgets.

\subsubsection{Search baselines}

\textbf{Random search} samples $\mathcal{A}$ uniformly; despite its simplicity it is a strong NAS baseline \citep{Li2020d}.
\textbf{Regularized evolution} \citep{Real2019a} mutates tournament winners from an aging population.
Both are matched to the RL arms on the number of \textit{real} evaluations consumed.

\subsection{Conformal evaluation filtering}
\label{ssec:filters}

\subsubsection{The filtering loop}
\label{sssec:loop}

After calibration on a warm-up set, each filter provides a reward upper bound $U(x)$ for candidate features $x=x(a)$.
It receives feedback $(x,r)$ after each real evaluation, where $r=R(a)$; the update rule depends on the filter.
The filtered search loop proceeds as follows:
\begin{enumerate}
    \item \textbf{Warm-up.} The first $W$ candidates are always evaluated (the filter is inactive), forming the initial calibration set.
    \item \textbf{Filtering.} Thereafter, a proposed candidate is evaluated if and only if $U(x) \ge \tau$, where $\tau$ is the user-chosen acceptance threshold on the reward; otherwise it is \textbf{pruned} (never trained) and the policy receives a fixed rejection reward $r_{\text{rej}} = -1$, which discourages the controller from proposing candidates the filter would prune.
    This value shapes the proposal distribution and interacts with the filter; we hold it fixed across all arms for comparability.
    \item \textbf{Feedback.} Each real evaluation $(x,r)$ is passed to the filter's update rule.
\end{enumerate}
A false discard occurs when $U(x)<\tau\le r$, which implies $r>U(x)$.
We therefore target a one-sided miscoverage rate $\delta\in(0,1)$.
The coverage statements below concern the stream on which feedback is observed; conditional false-discard rates are measured separately.
Empirical coverage is measured on the post-warm-up \textit{evaluated} stream as the fraction of candidates with $r \le U(x)$; the pruned fraction is the evaluation saving.
In the batched bandit runs, candidates are pruned using the offset held fixed at the start of each batch of $B = 20$, whereas this reported coverage uses the offset updated between accepted evaluations; these decision-time and update-time bounds coincide at $B = 1$, and the distinction is detailed in Appendix~\ref{app:proof}.
Our coverage statements thus concern the update-time bounds, not the bounds that drive pruning.
In the sequential MDP the filter gates each step's partial-network evaluation, and coverage is additionally reported \textit{conditional on layer depth}, where the within-episode distribution shift is strongest.
For uncensored measurement of depth-conditional coverage we use a \textbf{no-prune audit} mode in which every candidate is evaluated (pruning disabled) while the filter otherwise runs exactly as in a live search, so coverage is observable at depths that a realistic $\tau$ would otherwise censor.

\subsubsection{Reward surrogate}

The surrogate $g(x)$ is a three-layer multilayer perceptron (hidden width 64, ReLU), trained by mean-squared error, using the architecture of \citep{Fayyazi2026}.
Its input is the family-specific feature encoding $x(a)$: each categorical knob is one-hot encoded and the per-block bit-widths become one binary flag per block, zero-padded to the family's maximum depth, giving a fixed dimension $d$ per family (e.g.\ $d = 23$ for the ResNet-like family); the sequential space encodes per-layer slots for all possible layers plus cumulative-context features.
Whenever a filter is (re)calibrated, the available data is split: the surrogate is fit on the first half and nonconformity scores are computed on the held-out second half, following split (inductive) conformal prediction \citep{Papadopoulos2002}.
The held-out split avoids estimating prediction error from the surrogate's own training residuals, which can understate generalization error.

\subsubsection{Static split-conformal baselines}

Given held-out absolute residuals $\{|r_i - g(x_i)|\}_{i=1}^{n}$, the split conformal offset is their empirical $(1-\delta)$-quantile,
\begin{equation}
\label{eq:split}
\alpha = Q_{1-\delta}\big(\{\,|r_i - g(x_i)|\,\}_{i=1}^{n}\big),
\quad
U(x) = g(x) + \alpha .
\end{equation}
Under exchangeability, the finite-sample-corrected split-conformal quantile guarantees $\mathbb{P}\{r\le U(x)\}\ge1-\delta$ \citep{Vovk2022, Lei2018}.
We use the plain empirical quantile rather than the finite-sample-corrected level.
We evaluate two static variants:
\begin{itemize}
    \item \textbf{Calibrate-once} (\texttt{split}): $\alpha$ is fixed after warm-up.
    \item \textbf{Periodically recalibrated} (\texttt{split-recal}): every $P = 50$ evaluated candidates, the surrogate is re-fit and Eq.~\eqref{eq:split} is recomputed on all data observed so far (warm-up plus accepted candidates).
    Periodic recalibration is MARCO's stated mitigation of distribution shift; their paper specifies no period, so $P = 50$ is an inferred setting, documented here.
\end{itemize}
The exchangeability premise is precisely what RL search violates: the candidate distribution at step $t$ is the \textit{current policy's} proposal distribution, not the warm-up distribution, and it keeps moving as the policy improves.
In the sequential MDP the violation is compounded within each episode, since shallow and deep partial networks are drawn from visibly different distributions.
Periodic recalibration tracks this shift only with a lag and offers no guarantee between recalibrations.

\subsubsection{Adaptive conformal inference}

Adaptive Conformal Inference (ACI, \cite{Gibbs2021}) abandons the fixed quantile and \textit{controls} coverage online.
We use an additive-offset parametrization: after each real evaluation, with the miscoverage indicator $\text{err}_t = \mathbf{1}\{ r_t > g(x_t) + \alpha_t \}$, the offset is updated
\begin{equation}
\label{eq:aci}
\alpha_{t+1} = \alpha_t + \gamma\,\big(\text{err}_t - \delta\big),
\quad \gamma = 0.05,
\end{equation}
rising by $\gamma(1-\delta)$ on each miscoverage and decaying by $\gamma\delta$ otherwise, so increases and decreases balance when the average miscoverage equals $\delta$.
Equation~\eqref{eq:aci} is online subgradient descent on the pinball loss at level $1-\delta$ of the bound residual, and the standard ACI telescoping argument applies directly: summing Eq.~\eqref{eq:aci} over $t$ gives the signed identity $\frac{1}{T}\sum_{t}(\text{err}_t - \delta) = (\alpha_{T+1} - \alpha_1)/(\gamma T)$, so taking absolute values,
\begin{equation}
\label{eq:telescope}
\Big|\frac{1}{T}\sum_{t=1}^{T} \text{err}_t - \delta\Big|
= \frac{|\alpha_{T+1} - \alpha_1|}{\gamma\, T}
\le \frac{C}{\gamma\, T},
\end{equation}
where $C$ is finite for bounded residuals and hence bounded offsets; Appendix~\ref{app:proof} gives a self-contained derivation with an explicit constant and extends it to the normalized and Mondrian variants.
Long-run empirical coverage therefore converges to $1-\delta$ at rate $O(1/T)$ \textit{for an arbitrary bounded-residual sequence} under an uninterrupted recursion (no exchangeability, stationarity or distributional assumption), which is exactly the guarantee that survives RL-induced drift; being a long-run rate, the bound may be loose when few candidates are evaluated.
The step size is deliberately \textit{constant}: a decaying $\gamma$ would recover a static quantile in the limit and forfeit the ability to track persistent shift \citep{Angelopoulos2024}.
The offset is initialized from the split calibration of Eq.~\eqref{eq:split} at the end of warm-up.

\subsubsection{AgACI: aggregating step sizes}

ACI requires a step size $\gamma$.
To avoid choosing a single step size, we run $K$ ACI experts with $\gamma_k\in\{0.01,0.02,0.05,0.1,0.2\}$.
Each updates its offset $\alpha_t^{(k)}$ by Eq.~\eqref{eq:aci}.
Exponential weighting gives $\bar\alpha_t=\sum_k w_t^{(k)}\alpha_t^{(k)}$ and $U(x)=g(x)+\bar\alpha_t$.
After each evaluation, expert $k$ incurs the pinball loss of its own bound at level $1-\delta$,
\begin{equation*}
    \begin{aligned}
        \ell^{(k)}_t &= \ell_{1-\delta}\big(r_t - (g(x_t) + \alpha^{(k)}_t)\big),\\
        \ell_{1-\delta}(z) &= \begin{cases} (1-\delta)\,z & z \ge 0,\\ -\delta\,z & z < 0.\end{cases}
    \end{aligned}
\end{equation*}
We normalize the losses at each step as $\tilde\ell_t^{(k)}=(\ell_t^{(k)}-\min_j\ell_t^{(j)})/(\max_j\ell_t^{(j)}-\min_j\ell_t^{(j)})\in[0,1]$.
If all experts tie, weights remain unchanged.
Otherwise, Hedge \citep{Freund1997} uses the anytime learning rate $\eta_t=\sqrt{8\ln K/t}$ \citep{CesaBianchi2006}:
\begin{equation}
\label{eq:hedge}
w^{(k)}_{t+1} \propto w^{(k)}_t \exp\big(-\eta_t\, \tilde{\ell}^{(k)}_t\big).
\end{equation}
The grid, rate schedule, and loss are fixed a priori.
We use an AgACI-inspired aggregation scheme \citep{Zaffran2022} with additive-offset experts and Hedge weighting of normalized pinball losses (Eq.~\eqref{eq:hedge}), rather than the original Bernstein Online Aggregation procedure.
We retain the code label \texttt{agaci}, but do not claim the original method's formal guarantees.

\subsubsection{Normalized conformal: a per-candidate width}
\label{sssec:normalized}

A global offset $U(x)=g(x)+\alpha$ ranks candidates exactly as $g(x)$ does, providing no candidate-dependent exploration signal.
Its marginal coverage may also conceal under-coverage in harder sub-populations.
The locally-adaptive (normalized) filter \citep{Papadopoulos2008, Lei2018} divides the nonconformity score by a per-candidate difficulty scale $u(x) > 0$:
\begin{equation}
\label{eq:normalized}
    \begin{aligned}
        s_i &= \frac{|r_i - g(x_i)|}{u(x_i)},\\
        q &= Q_{1-\delta}\big(\{s_i\}\big),\\
        U(x) &= g(x) + q\,u(x),
    \end{aligned}
\end{equation}
with the ACI recursion of Eq.~\eqref{eq:aci} now applied to the multiplier $q$, driven by the normalized miscoverage event $r_t > g(x_t) + q_t\,u(x_t)$.
The difficulty scale is supplied by a \emph{mean-absolute-deviation} (MAD) surrogate: a second network $h(x)$, of the same architecture as $g$, is fit to the mean network's absolute residuals $|r - g(x)|$, and
\begin{equation}
\label{eq:mad}
u(x) = \max\big(h(x),\; \varepsilon_u\big), \quad \varepsilon_u = 10^{-3},
\end{equation}
where the floor guards against degenerate scales.
Coverage remains marginal.
The candidate-dependent width $q\,u(x)$ supplies the exploration signal used in Sec.~\ref{ssec:acquisition}.

\subsubsection{Mondrian conformal: group-conditional coverage}

For coverage conditional on construction depth, we use a Mondrian construction \citep{Vovk2022, Bostroem2020} with one ACI offset per group $k=\mathrm{group}(x)$:
\begin{equation}
\label{eq:mondrian}
\begin{aligned}
U(x) &= g(x) + \alpha_{\text{group}(x)},\\
\alpha_{k,\,t+1} &= \alpha_{k,t} + \gamma\big(\mathbf{1}\{r_t > g(x_t) + \alpha_{k,t}\} - \delta\big),
\end{aligned}
\end{equation}
updating only the observed candidate's group.
At calibration, $\alpha_k$ is initialized from the group's own residual quantile when the group has at least $n_{\min} = 10$ calibration points, and from the global offset of Eq.~\eqref{eq:split} otherwise.
After warm-up, each group adapts by its own online ACI recursion from its first accepted observation onward, so $n_{\min}$ governs only initialization, not continued adaptation.
Each sufficiently-observed group then inherits the guarantee of Eq.~\eqref{eq:telescope} \textit{within the group}, at the price of slower adaptation (each offset learns only from its own group's stream) and of the explicit small-group fallback.

\subsubsection{Gaussian-process baseline}

We compare conformal calibration with a GP bound at the same nominal coverage level.
The GP uses a radial-basis-function (RBF) plus white-noise kernel, fits hyperparameters by marginal likelihood, and is re-fit online on all evaluated data.
It prunes using the upper bound \citep{Srinivas2010}
\begin{equation}
\label{eq:gpucb}
U(x) = \mu(x) + \kappa\,\sigma(x), \quad \kappa = \Phi^{-1}(1-\delta).
\end{equation}
Here $\mu(x)$ and $\sigma(x)$ are the GP posterior mean and standard deviation, and $\Phi$ is the standard normal cumulative distribution function.
The one-sided multiplier is $\kappa\approx1.28$ at $\delta=0.1$.
This bound depends on the GP model assumptions and has no distribution-free coverage guarantee.
Its multiplier does not adapt to observed miscoverage.
ACI instead updates its offset from coverage errors according to Eq.~\eqref{eq:aci}.

\subsubsection{Pluggable uncertainty: calibrating a GP or ensemble width}
\label{sssec:pluggable}

Any candidate-dependent uncertainty estimate can supply $u(x)$ in Eq.~\eqref{eq:normalized}, with ACI calibrating its multiplier online.
The Gaussian-process variant, denoted \texttt{gp-aci}, uses the GP posterior mean as $g(x)$ and $u(x)=\max(\sigma(x),\varepsilon_u)$.
It updates $q$ by Eq.~\eqref{eq:aci} instead of fixing the multiplier at $\kappa$.
The deep-ensemble variant, denoted \texttt{ensemble-aci}, uses five independently initialized networks with the architecture of the MLP surrogate.
Their mean predicts the reward, and their prediction standard deviation, floored at $\varepsilon_u$, supplies $u(x)$.
Both use the normalized ACI update unchanged.
Section~\ref{ssec:e3} compares these uncertainty sources with the learned MAD scale and fixed GP bound.

\subsection{Summary of coverage guarantees}
\label{ssec:guarantees}

The filters provide coverage statements with different assumptions and conditioning.
Online statements concern the observed feedback stream and require bounded residuals (normalized residuals for normalized ACI):
\begin{enumerate}
    \item \textbf{Split conformal, Eq.~\eqref{eq:split}:} finite-sample marginal coverage under exchangeability when the corrected quantile is used; our empirical-quantile implementation approximates this construction.
    RL-driven search violates the exchangeability premise.
    Periodic recalibration mitigates the violation but lags the drift and provides no guarantee between recalibrations.
    \item \textbf{ACI and its normalized form, Eqs.~\eqref{eq:aci} and~\eqref{eq:normalized}:} distribution-free \textit{long-run} coverage, Eq.~\eqref{eq:telescope}, valid for arbitrary bounded-residual sequences under an uninterrupted recursion, including policy-improvement drift and the within-episode depth shift of the sequential MDP.
    The guarantee is marginal over the stream and holds for any single online offset, whether global (ACI) or locally scaled by the $u(x)$ of Eq.~\eqref{eq:mad} (normalized).
    Each AgACI expert has this property, but their weighted-average offset is not an ACI iterate and does not inherit Eq.~\eqref{eq:telescope}; the aggregate instead relies on expert weighting by normalized pinball loss, and its empirical calibration is reported in Sec.~\ref{sec:results}.
    \item \textbf{Mondrian ACI, Eq.~\eqref{eq:mondrian}:} the guarantee of Eq.~\eqref{eq:telescope} \textit{per group} for groups with sufficient observations; sparse groups initialize from the global offset and subsequently adapt from their own observations.
\end{enumerate}
The GP bound, Eq.~\eqref{eq:gpucb}, has no distribution-free coverage guarantee; and since we use it here as an \emph{acceptance criterion} rather than an acquisition function, GP-UCB's optimization-regret guarantees do not apply to this setting.
In acquisition search (Sec.~\ref{ssec:acquisition}), the same feedback update calibrates the bound on evaluated candidates; coverage alone does not imply a search-regret guarantee.

\subsection{Conformal acquisition: from pruning to directed search}
\label{ssec:acquisition}

The filters of Sec.~\ref{ssec:filters} \textit{gate} evaluations proposed by an external controller.
Acquisition instead selects candidates using the normalized bound $U(x)=g(x)+q\,u(x)$, combining predicted reward with candidate-dependent optimism.
Three proposal mechanisms share a single loop: a warm-up of $W$ uniformly-sampled evaluations calibrates the filter, after which each step proposes one candidate, evaluates it, and applies the online ACI update; the surrogate is fixed after warm-up and the offset recursion is uninterrupted, so the coverage guarantee of Eq.~\eqref{eq:telescope} holds on the evaluated stream.
The three mechanisms share this calibration-state update procedure, but differ in proposal generation (greedy pool-argmax, MCTS, or the optimism-shaped controller), controller training, and evaluation granularity: the shaped controller in particular uses a different training reward and partial-network evaluations.
Appendix~\ref{app:algos} gives the consolidated pseudocode.

\textbf{(a) Greedy acquisition.}
Greedy acquisition maximizes the bound for the next evaluation without lookahead.
Its uncertainty term $q\,u(x)$ adds an exploration bonus when $q > 0$; the online update can drive $q$ negative, in which case the term penalizes rather than rewards uncertainty.
Each step draws a cheap pool of $S = 64$ candidates uniformly from the space and evaluates the pool's bound-argmax:
\begin{equation}
\label{eq:greedy}
a_{\text{next}} = \arg\max_{a \in \text{pool}} \; U\big(x(a)\big).
\end{equation}
This tests whether calibrated optimism can direct search without a learned policy or tree.

\textbf{(b) Optimism-shaped reinforcement learning.}
We also train the sequential controller of Sec.~\ref{sssec:sequential} with a reward shaped toward finding better architectures.
PPO optimizes expected proposal reward, while architecture search is judged by the best candidate found.
Reward shaping tests whether better alignment improves search.
The shaped per-step training reward combines improvement over the incumbent (an expected-improvement analogue) with an intrinsic optimism bonus:
\begin{equation}
\label{eq:shaped}
\tilde{r}_t = \max\big(0,\; r_t - r^{\text{best}}_{<t}\big) + \beta\, q\,u(x_t),
\quad \beta = 0.1,
\end{equation}
where $r^{\text{best}}_{<t}$ is the best raw reward observed before step $t$; incumbent tracking and all reported metrics use the raw reward.
As Sec.~\ref{ssec:e6} reports, this shaping does not reliably improve on the unshaped controller and can destabilize it; we include it as an exploratory negative result, not as a recommended proposer.

\textbf{(c) Conformal-optimistic tree search.}
In the sequential space the construction process is itself a tree: a node is a partial network, an edge is the next layer's choice, and a leaf is a complete architecture.
An AlphaZero-style planner \citep{Silver2018} selects each proposal.
The tree is descended by the PUCT rule
\begin{equation}
\label{eq:puct}
a = \arg\max_{a'} \; Q(s,a') + c_{\text{puct}} \, P(a' \mid s)\,
\frac{\sqrt{\sum_{b} N(s,b) + 1}}{1 + N(s,a')},
\end{equation}
with $c_{\text{puct}} = 1.4$, where $Q$ is the mean backed-up value, $N$ the visit count and $P$ the policy prior.
The $+1$ inside the square root is a deliberate modification of the standard PUCT numerator $\sqrt{\sum_b N(s,b)}$ that keeps the exploration term positive before any child of $s$ is visited.
Each simulation completes a partial network and backs up its calibrated bound $U$, without training the completion.
Thus, simulations consume no real evaluations.
After $N_{\text{sim}} = 200$ simulations the visit-count-greedy leaf is committed and receives the single real evaluation of that step; each real evaluation applies the online ACI update to the calibration state (the surrogate is fixed after warm-up), closing the loop under drift.
Tree search for NAS is not itself new (AlphaX, \cite{Wang2020b}); the novel element is the calibrated-optimistic leaf value with conformally-controlled coverage on a policy-drifting stream.

Greedy acquisition, MCTS, and the random, evolution, and PPO baselines are compared at matched numbers of complete-network evaluations.
The shaped controller uses per-step evaluations and is compared separately with its unshaped counterpart at matched episode counts.

\subsection{Evaluation protocol}
\label{ssec:protocol}

\begin{itemize}
    \item \textbf{Coverage.} The fraction of post-warm-up evaluated candidates with $r \le U(x)$; reported marginally and, in the sequential MDP, conditional on layer depth (measured in the no-prune audit mode of Sec.~\ref{sssec:loop} to avoid censoring).
    \item \textbf{Calibration error and steerability.} $|\text{coverage} - (1-\delta)|$, swept over $\delta \in \{0.05, 0.1, 0.2, 0.3\}$.
    A filter is \textit{steerable} if its realized coverage moves monotonically with the requested level.
    \item \textbf{Evaluation saving and search efficiency.} The fraction of proposed candidates pruned in filtered search, or the best reward reached at matched evaluation budgets in directed search.
    \item \textbf{Solution quality.} The best accuracy and reward discovered, compared across filters at fixed seed and acceptance threshold.
    \item \textbf{False discards.} We use two distinct metrics with different denominators.
    CNN replay (Sec.~\ref{ssec:e2}) reports the \emph{discarded-good share}, the fraction of pruned candidates with $r\ge\tau$.
    The tabular benchmark (Sec.~\ref{ssec:e3}) reports the \emph{false-discard rate}, the fraction of good post-warm-up cells ($r_{200}\ge\tau_{\mathrm{ref}}$) that are pruned, and pruning power among cells with $r_{200}<\tau_{\mathrm{ref}}$.
    Pruning uses the 12-epoch upper bound and a separate threshold $\tau_{\mathrm{proxy}}$.
    These conditional rates have different denominators and are not bounded by evaluated-stream coverage alone.
    In the tabular experiment, the link between false discards and miscoverage is further limited by the different fidelities used for pruning and scoring.
    \item \textbf{Reproducibility.} Across-seed standard deviation of coverage (three seeds for the calibration sweeps; eight for the depth-conditional audits, whose per-depth estimates are far noisier because each depth group holds only a fraction of the stream).
\end{itemize}

Default settings are summarized in Table~\ref{tab:settings}.

\section{Experiments}
\label{sec:results}
We evaluate \textbf{(E1)} calibration, steerability and reproducibility; \textbf{(E2)} solution quality and false discards; \textbf{(E3)} false discards under a realistic low-fidelity proxy with measured hardware costs; \textbf{(E4)} depth-conditional coverage; \textbf{(E5)} controller efficiency; and \textbf{(E6)} directed search.

\subsection{Experimental setup}

\textbf{Benchmark grid.}
Three architecture families, each paired with its standard dataset: a LeNet-like family on MNIST, a ResNet-like family on CIFAR-10, and a depthwise-separable MobileNet-like family on CIFAR-100 (search spaces of Sec.~\ref{sssec:spaces}; per-block bit-widths $\{4, 8\}$ throughout, applied by fake post-training quantization).

\textbf{Search and budgets.}
Bandit experiments run the CTDE-PPO controller for 50 iterations $\times$ 20 trials $=$ 1{,}000 proposed candidates, the first 100 forming the warm-up/calibration set.
Prune fractions from live filtered CNN runs are reported over all 1{,}000 proposed candidates (the warm-up 100 are always evaluated); the offline CNN replay (Table~\ref{tab:fd}) and the tabular benchmark (Sec.~\ref{ssec:e3}) instead report post-warm-up fractions.
These denominators are not directly comparable.
Sequential experiments run 150 construction episodes with per-step evaluation.
The reward, Eq.~\eqref{eq:reward}, uses $\lambda = 0.2$, $\nu = 1.0$, with resource-constrained budgets ($M_B = 512$\,kB; $T_B = 0.3$\,ms), tightened to $M_B = 256$\,kB for the directed-search experiments (Sec.~\ref{ssec:e6}).
We select $\tau$ from an unfiltered probe run for each family: 0.35 for ResNet-like, 0.89 for LeNet-like (between its two reward modes), and 0.18 for MobileNet-like.
These thresholds reflect each family's reward distribution and are not transferable constants.

\textbf{Evaluation feedback.}
All experiments use test-split accuracy as search feedback: partial-training accuracy for the CNN families and the tabulated 12-epoch accuracy for NAS-Bench-201.
On NAS-Bench-201 the 200-epoch accuracy is used only to score false discards retrospectively, not as search feedback; the synthetic-noise diagnostic of Appendix~\ref{app:noise} instead perturbs the 200-epoch reward.
We use no separate validation holdout, although NAS-Bench-201 provides validation measurements.
Because the same split guides search and scores its outcomes, the reported accuracies do not provide an independent estimate of generalization.

\textbf{Filter arms.}
\texttt{adaptive} (ACI, $\gamma = 0.05$), \texttt{agaci} (aggregation over a fixed step-size grid, avoiding a single $\gamma$ but still requiring the grid and an expert-weighting schedule), \texttt{split} (calibrate-once split conformal), \texttt{split-recal} (periodically recalibrated split conformal), \texttt{gpbo} (GP-UCB at the same nominal level), and \texttt{none} (no filtering).
All arms share the warm-up data and controller, and use the same code version and seeds within an experiment.
The MLP-based filters share a surrogate architecture; GP-UCB uses the GP described in Sec.~\ref{ssec:filters}.
These are the arms of the calibration sweep.
Three further conformal variants defined in Sec.~\ref{sec:methods}, normalized ACI (\texttt{normalized}), GP-backed ACI (\texttt{gp-aci}) and ensemble-backed ACI (\texttt{ensemble-aci}), are evaluated on the tabular benchmark (Sec.~\ref{ssec:e3}) and in Appendix~\ref{app:noise}, and depth-conditional Mondrian ACI (\texttt{mondrian}) is used in the depth audit (Sec.~\ref{ssec:e4}).

\subsection{E1: Calibration, steerability and reproducibility}
\label{ssec:e1}

\begin{figure*}[t]
\centering
\includegraphics[width=0.95\textwidth]{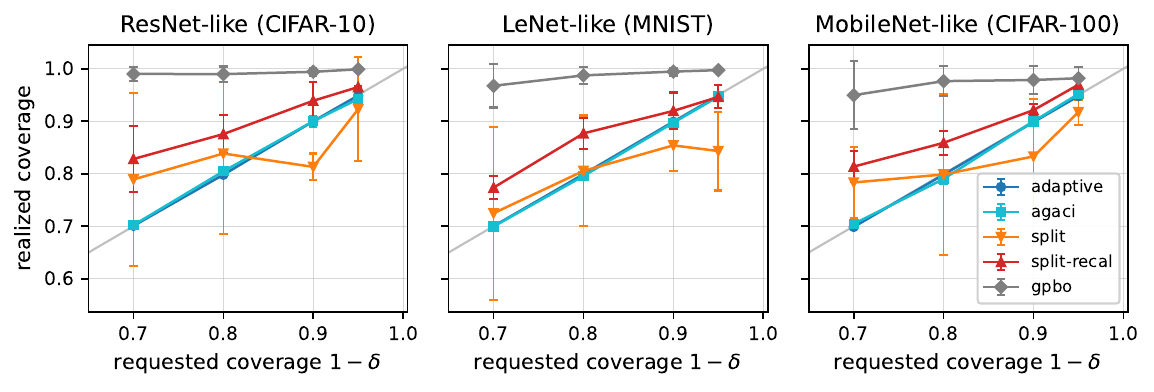}
\caption{
    Realized coverage vs requested target $1-\delta$, per filter arm and family (points = seed mean, error bars = across-seed standard deviation; grey line = ideal calibration).
    ACI and AgACI track the target with small across-seed deviations.
    GP-UCB remains near 0.99 across requested levels.
    Calibrate-once split under-covers at tight targets and over-covers at loose ones, with substantially larger across-seed variation.
}
\label{fig:calibration}
\end{figure*}

E1 checks cumulative coverage tracking on the evaluated stream, as anticipated by the telescoping identity (Eq.~\eqref{eq:telescope}).
Coverage is measured using update-time bounds, which can differ from the pre-batch bounds used for pruning (Sec.~\ref{sec:methods}).
Sections~\ref{ssec:e2}--\ref{ssec:e3} assess the consequences for pruning and search quality.
Table~\ref{tab:cal-cifar10} reports the five-arm $\delta$-sweep on ResNet-like/CIFAR-10; the per-target coverages behind it are in Table~\ref{tab:cov-cifar10}.
Figure~\ref{fig:calibration} extends the comparison to all three families.

\begin{table}[t]
\centering
\caption{Calibration error $|\text{coverage} - (1-\delta)|$ over the $\delta$-sweep, ResNet-like/CIFAR-10.}
\label{tab:cal-cifar10}
\footnotesize
\setlength{\tabcolsep}{3pt}
\begin{tabular}{@{}lcclc@{}}
\toprule
arm & mean $|$err$|$ & max $|$err$|$ & steerable? & max std \\
\midrule
\textbf{adaptive} & \textbf{0.001} & 0.002 & monotone & \textbf{0.002} \\
\textbf{agaci} & \textbf{0.004} & 0.008 & monotone & 0.011 \\
split & 0.061 & 0.089 & \textbf{non-monotone} & \textbf{0.165} \\
split-recal & 0.065 & 0.128 & over-covers, drifts & 0.063 \\
gpbo & 0.156 & 0.290 & \textbf{fixed $\approx$ 0.99} & 0.010 \\
\bottomrule
\end{tabular}
\end{table}

\begin{table}[t]
\centering
\caption{Realized coverage on ResNet-like/CIFAR-10 with the weak proxy; entries are means over three seeds.}
\label{tab:cov-cifar10}
\footnotesize
\begin{tabular}{@{}lcccc@{}}
\toprule
& \multicolumn{4}{c}{Requested coverage $1-\delta$} \\
\cmidrule(lr){2-5}
arm & 0.95 & 0.90 & 0.80 & 0.70 \\
\midrule
adaptive & 0.949 & 0.900 & 0.798 & 0.701 \\
agaci & 0.942 & 0.900 & 0.805 & 0.702 \\
split & 0.923 & 0.813 & 0.839 & 0.789 \\
split-recal & 0.965 & 0.939 & 0.876 & 0.828 \\
gpbo & 0.999 & 0.994 & 0.990 & 0.990 \\
\bottomrule
\end{tabular}
\end{table}

\textbf{ResNet-like/CIFAR-10 (weak proxy, $\tau = 0.35$, seeds 0/1/2).}
ACI achieves a mean calibration error of 0.001 and maximum error of 0.002, with monotone response to the target and maximum across-seed standard deviation of 0.002 (Table~\ref{tab:cal-cifar10}).
AgACI achieves a mean error of 0.004 without selecting a single step size.
The baselines show three different limitations.
Calibrate-once split under-covers at tight targets and over-covers at loose ones, with seed standard deviation up to 0.165.
At $\delta=0.2$, coverage ranges from 0.61 to 0.98 across seeds, spanning substantial under- and over-coverage.
Periodic recalibration reduces this variability but retains systematic over-coverage: lowering the target from 0.95 to 0.70 changes coverage from 0.965 to 0.828, leaving an error of 0.128 at the loosest target.
GP-UCB coverage remains near 0.99 with little seed variation, while its prune rate changes only from 0.36 to 0.42 across the sweep.

\textbf{Replication across families} (Figure~\ref{fig:calibration}).
The ranking also holds on LeNet-like/MNIST ($\tau=0.89$) and MobileNet-like/CIFAR-100 ($\tau=0.18$, strong proxy).
Mean calibration errors are 0.001 and 0.002 for ACI, and 0.003 and 0.004 for AgACI, respectively; ACI's seed standard deviation is at most 0.001.
GP-UCB has mean errors of 0.149 and 0.134, with coverage near 0.95--0.99 and prune rates largely insensitive to $\delta$.
At target 0.70, periodically recalibrated split remains at 0.774 and 0.814.
Calibrate-once split again under-covers at tight targets and over-covers at loose ones, with seed standard deviations up to 0.165 on LeNet-like and 0.153 on MobileNet-like.

\begin{figure}[t]
\centering
\includegraphics[width=0.65\columnwidth]{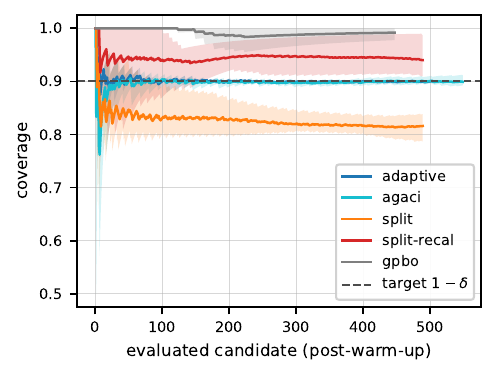}
\caption{
    Cumulative coverage over post-warm-up evaluated candidates (ResNet-like/CIFAR-10, weak proxy, $\delta=0.1$, target coverage 0.90).
    Lines show means over three seeds; shaded bands show the min--max range.
    ACI and AgACI track the target, while the split-conformal baselines and GP-UCB settle off-target.
    Because each filter prunes a different number of candidates, the curves end at different $x$-values, one per arm's count of evaluated candidates.
}
\label{fig:runcov}
\end{figure}

\textbf{Coverage dynamics under drift.}
Running coverage (Fig.~\ref{fig:runcov}) shows how these differences develop over time.
ACI converges from initial over-coverage to the target and remains close under policy drift, with little across-seed variation.
The static bounds settle off-target: periodically recalibrated split over-covers with a seed range of approximately 0.91--0.99, while calibrate-once split eventually under-covers.

\textbf{Evaluation savings.}
For ACI, increasing the tolerated miscoverage $\delta$ increases evaluation savings.
On the LeNet-like family the adaptive arm prunes 30.3\% of candidates at $\delta = 0.05$ rising monotonically to 39.5\% at $\delta = 0.3$; on the ResNet-like family, 26.8\% rising to 36.3\%.
On the MobileNet-like family the trend is the same but noisier across seeds: 34.6\% at $\delta = 0.05$ rising to 46.5\% at $\delta = 0.2$, then 43.1\% at $\delta = 0.3$.
The baseline prune rates respond less consistently to the target.
Periodically recalibrated split over-covers at loose targets.
GP-UCB's prune rate changes little with $\delta$: approximately 0.33 on LeNet-like and 0.56--0.58 on MobileNet-like.

\textbf{Acceptance threshold at fixed miscoverage.}
At fixed $\delta=0.1$, increasing $\tau$ changes ACI's savings while leaving evaluated-stream coverage nearly unchanged (ResNet-like, three seeds).
Across thresholds within the main reward distribution, prune rates rise from 0.200 to 0.262, 0.352, and 0.486, while coverage stays at 0.899--0.901 with seed standard deviation at most 0.002.
The prune rates at the calibrated operating points are consistent with the 22--32\% reported in prior work \citep{Fayyazi2026}.
Periodically recalibrated split does not separate the two effects: its measured coverage rises from 0.897 to 0.981 as the acceptance threshold increases and changes the evaluated population.
At $\tau=0.45$, in the upper tail of the reward distribution, pruning reaches approximately 73\% and coverage control weakens for all arms as evaluations become sparse.
Because this high threshold restricts the accessible population, the best architecture reached in this separate threshold sweep is 0.497 (below the 0.505 optimum of the operating-threshold runs); ACI and AgACI fall slightly further, to 0.489 and 0.482.
The looser static bounds retain the peak while pruning less.
We therefore select operating thresholds within the main reward distribution.

\subsection{E2: Solution quality and false discards}
\label{ssec:e2}

At the family-specific operating thresholds, ACI and AgACI discover the same best architecture as the unfiltered baseline across the tested $\delta$ values, seeds, and families.
Pruning 25--50\% of evaluations causes no measured loss in solution quality at these settings.
In the ResNet-like/CIFAR-10 five-arm $\delta$-sweep (weak proxy, $\tau = 0.35$, three seeds), periodically recalibrated split returns a worse optimum: best accuracy is 0.481, compared with 0.505 for every other arm, uniformly across seeds and $\delta$.
The per-trial logs attribute this to its recalibrated bound pruning the top-scoring architecture before it is evaluated.

The observed accuracy loss is largest where the optimum is isolated: periodically recalibrated split falls 0.024 below the other arms on ResNet-like.
Near-optimal alternatives limit the loss on LeNet-like (an accuracy plateau near 0.98; penalty approximately 0.002) and MobileNet-like (penalty 0).

\textbf{Measured discarded-good shares.}
We also replay each filter over a fully evaluated, unfiltered stream, rebuilding features from logged configurations.
Each filter is calibrated on the warm-up prefix and updated only on candidates it would accept.
This is a sequential, off-policy replay that decides and updates one candidate at a time, so it does not reconstruct the live batch decisions (which commit a batch of 20 before any update); it retains the rewards of pruned candidates for analysis.
Here the discarded-good share (distinct from the full-fidelity false-discard rate of Sec.~\ref{ssec:e3}) is the fraction of pruned candidates whose reward reaches $\tau$ (Table~\ref{tab:fd}).
For ACI and AgACI, this share generally rises with $\delta$.
Calibrate-once split responds less consistently: on LeNet-like it reaches 0.11 at $\delta=0.05$ and then plateaus near 0.20.
Its coverage falls from 0.77 to 0.66 while it prunes approximately 40\% of proposals.
On MobileNet-like/CIFAR-100 with the strong proxy, discarded-good shares rise to approximately 0.07 for ACI and AgACI, while calibrate-once split plateaus near 0.08 and under-covers.
GP-UCB retains coverage near 1.0.

Coverage calibration does not necessarily minimize discarded-good candidates.
In these replays, ACI and AgACI show a more consistent response to $\delta$, whereas calibrate-once split has across-seed spreads up to $\pm0.10$ in discarded-good share and $\pm0.16$ in coverage, compared with approximately $\pm0.001$ coverage variation for ACI and AgACI.
Low discarded-good shares must be interpreted alongside pruning rates.
On ResNet-like at $\delta=0.05$, GP-UCB (fixed multiplier, online-refit GP) prunes 52.6\% of candidates with a discarded-good share of 0.0\%, compared with 26.0\% and 6.6\% for ACI.
Periodically recalibrated split also prunes more candidates than ACI with a lower discarded-good share.
These results show that overcoverage can coexist with substantial pruning and low discarded-good shares.
ACI provides explicit control of evaluated-stream coverage, but this does not necessarily yield a better pruning-discard trade-off.
Replay is off-policy and uses three seeds.
At $\delta=0.3$, periodically recalibrated split and GP-UCB each prune the optimum on one seed.
This supports the possibility of losing an optimum but differs from the live-search result above.

\begin{table*}[t]
\centering
\caption{
    Prune rate, discarded-good share, and realized coverage against the requested $\delta \in \{0.05, 0.10, 0.20, 0.30\}$ (offline replay over the no-prune streams, three-seed means).
    Prune rate is over post-warm-up candidates (the replay decides only after the warm-up prefix), so it is not directly comparable to the whole-run prune fractions of the live CNN runs.
    The \emph{discarded-good share} is the fraction of pruned candidates whose true reward reached $\tau$; its denominator is pruned candidates, distinct from the tabular false-discard rate of Sec.~\ref{ssec:e3}, whose denominator is good candidates.
    The first two families use the weak proxy; MobileNet-like/CIFAR-100 uses its strong proxy.
    ACI and AgACI track the coverage target while their discarded-good shares generally rise with $\delta$.
    Calibrate-once split under-covers and reaches a discarded-good share near 0.20 on MNIST; GP-UCB over-covers, with coverage near 1.0 and little prune-rate response to $\delta$.
}
\label{tab:fd}
\footnotesize
\setlength{\tabcolsep}{4pt}
\begin{tabular}{@{}l cccc cccc cccc@{}}
\toprule
& \multicolumn{4}{c}{prune rate} & \multicolumn{4}{c}{discarded-good share} & \multicolumn{4}{c}{coverage} \\
\cmidrule(lr){2-5} \cmidrule(lr){6-9} \cmidrule(lr){10-13}
arm & .05 & .10 & .20 & .30 & .05 & .10 & .20 & .30 & .05 & .10 & .20 & .30 \\
\midrule
\multicolumn{13}{@{}l}{\textit{ResNet-like / CIFAR-10} ($\tau = 0.35$, weak proxy)} \\
adaptive    & 0.260 & 0.369 & 0.488 & 0.519 & 0.066 & 0.090 & 0.176 & 0.194 & 0.950 & 0.900 & 0.800 & 0.700 \\
agaci       & 0.261 & 0.406 & 0.452 & 0.491 & 0.072 & 0.106 & 0.121 & 0.142 & 0.949 & 0.900 & 0.800 & 0.699 \\
split       & 0.253 & 0.373 & 0.385 & 0.450 & 0.067 & 0.100 & 0.099 & 0.099 & 0.973 & 0.867 & 0.831 & 0.738 \\
split-recal & 0.370 & 0.423 & 0.510 & 0.579 & 0.048 & 0.087 & 0.113 & 0.149 & 0.954 & 0.903 & 0.815 & 0.765 \\
gpbo        & 0.526 & 0.534 & 0.547 & 0.564 & 0.000 & 0.019 & 0.037 & 0.063 & 1.000 & 1.000 & 0.999 & 0.999 \\
\midrule
\multicolumn{13}{@{}l}{\textit{LeNet-like / MNIST} ($\tau = 0.89$, weak proxy)} \\
adaptive    & 0.319 & 0.366 & 0.380 & 0.387 & 0.008 & 0.046 & 0.078 & 0.104 & 0.949 & 0.899 & 0.800 & 0.700 \\
agaci       & 0.362 & 0.373 & 0.377 & 0.379 & 0.049 & 0.052 & 0.061 & 0.069 & 0.950 & 0.898 & 0.798 & 0.698 \\
split       & 0.402 & 0.432 & 0.440 & 0.440 & 0.114 & 0.190 & 0.204 & 0.204 & 0.774 & 0.752 & 0.731 & 0.661 \\
split-recal & 0.352 & 0.374 & 0.377 & 0.375 & 0.041 & 0.059 & 0.058 & 0.055 & 0.953 & 0.924 & 0.880 & 0.820 \\
gpbo        & 0.351 & 0.351 & 0.351 & 0.351 & 0.000 & 0.000 & 0.000 & 0.000 & 0.995 & 0.995 & 0.994 & 0.985 \\
\midrule
\multicolumn{13}{@{}l}{\textit{MobileNet-like / CIFAR-100} ($\tau = 0.18$, strong proxy)} \\
adaptive    & 0.341 & 0.395 & 0.450 & 0.491 & 0.006 & 0.013 & 0.037 & 0.068 & 0.949 & 0.899 & 0.799 & 0.706 \\
agaci       & 0.391 & 0.448 & 0.489 & 0.539 & 0.001 & 0.001 & 0.019 & 0.065 & 0.949 & 0.899 & 0.797 & 0.699 \\
split       & 0.489 & 0.576 & 0.581 & 0.581 & 0.040 & 0.082 & 0.082 & 0.082 & 0.770 & 0.790 & 0.741 & 0.741 \\
split-recal & 0.478 & 0.519 & 0.588 & 0.606 & 0.006 & 0.012 & 0.053 & 0.067 & 0.940 & 0.903 & 0.848 & 0.802 \\
gpbo        & 0.583 & 0.584 & 0.593 & 0.590 & 0.021 & 0.021 & 0.035 & 0.020 & 0.999 & 0.999 & 0.998 & 0.998 \\
\bottomrule
\end{tabular}
\end{table*}

\subsection{E3: Tabular benchmark under a realistic proxy}
\label{ssec:e3}

E3 asks whether coverage calibrated on low-fidelity rewards translates into retaining candidates that are good at full fidelity.
We complement the CNN replay measurements (Table~\ref{tab:fd}) with a tabular benchmark using measured hardware costs and a realistic low-fidelity proxy; a synthetic-noise diagnostic appears in Appendix~\ref{app:noise}.
The search observes NATS-Bench's 12-epoch CIFAR-100 rewards, while pruning decisions are scored against the 200-epoch reward median, $\tau_{\mathrm{ref}} = 0.438$.
We test $\delta \in \{0.05, 0.1, 0.2, 0.3, 0.4\}$ over $20$ seeds (setup in Table~\ref{tab:settings}; per-filter gaps to the raw-proxy ranking reference in Table~\ref{tab:nb201gap}).

Reusing the full-fidelity threshold for pruning yields post-warm-up false-discard rates of $0.83$--$0.99$ for the global filters.
The large reward shift across fidelities makes this threshold overly aggressive: 12-epoch rewards are on average $0.21$ lower (across-cell standard deviation $0.06$), despite strong rank correlation (Spearman $0.874$).
At 12 epochs, $99.1\%$ of good cells fall below $\tau_{\mathrm{ref}}$.
Using the \emph{retrospective}, benchmark-wide 12-epoch reward median ($0.214$), a quantity not available during a live search, as the pruning threshold instead keeps mean false-discard rates below $\delta$ for every filter, with similar best rewards but few evaluations saved.
Estimating this median from warm-up proposals would avoid querying the entire benchmark, but we do not evaluate that variant.

\begin{figure*}[t]
\centering
\includegraphics[width=\textwidth]{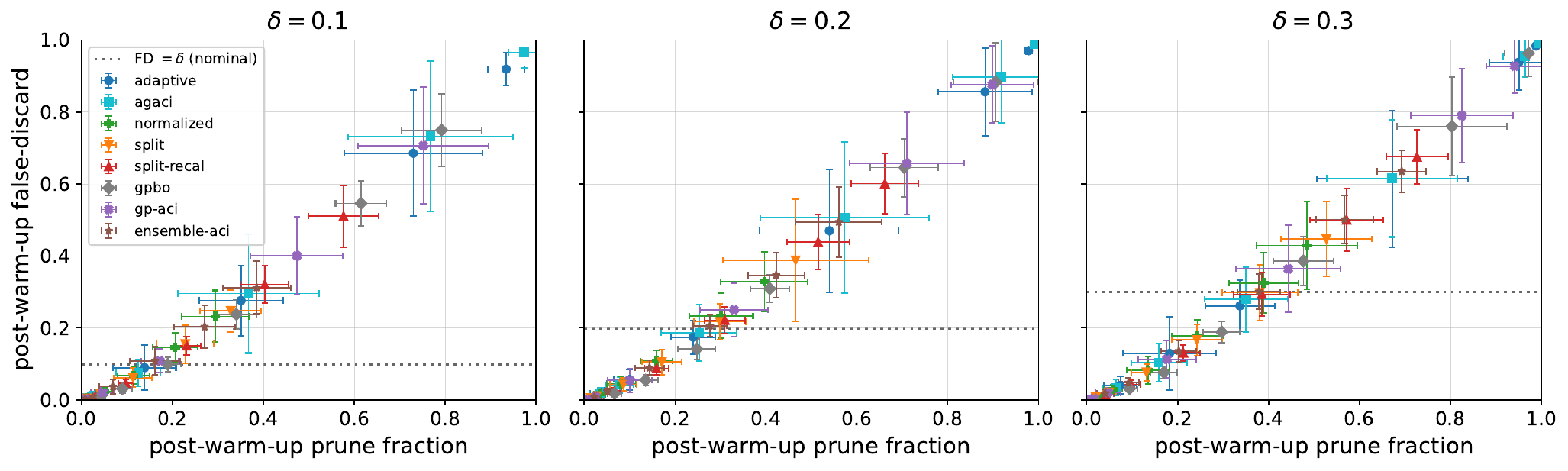}
\caption{
    Post-warm-up false-discard rate vs pruning fraction on NAS-Bench-201/CIFAR-100 under the 12-epoch proxy, scored against 200-epoch rewards ($\tau_{\mathrm{ref}} = 0.438$).
    Panels show three representative $\delta$ values.
    Each marker represents one filter and pruning threshold, swept from $\tau_{\mathrm{proxy}} = 0.214$ to $0.438$; points are seed means and bars are standard deviations on both axes ($20$ seeds).
    The dotted line marks $\mathrm{FD} = \delta$, a nominal comparison rather than a guaranteed bound.
    Paired comparisons with raw-proxy ranking appear in Fig.~\ref{fig:nb201gap} and Table~\ref{tab:nb201gap}.
}
\label{fig:nb201tradeoff}
\end{figure*}

Sweeping the pruning threshold reveals the cross-fidelity trade-off (Figure~\ref{fig:nb201tradeoff}): low false-discard rates require conservative pruning, and rates reach approximately $0.3$--$0.4$ at $40\%$ pruning.
The tested filters exhibit broadly similar trade-offs across all $\delta$.
At matched pruning fractions, their false-discard rates average $2.6$ percentage points above a retrospective raw-proxy ranking reference, which has access to every candidate's 12-epoch reward (Table~\ref{tab:nb201gap}).
No filter shows a consistent advantage over this reference.
Threshold selection strongly affects the observed trade-off, and observed-stream coverage alone does not determine full-fidelity pruning safety.

\subsection{E4: Depth-conditional coverage in sequential search}
\label{ssec:e4}

\begin{figure*}[t]
\centering
\includegraphics[width=0.95\textwidth]{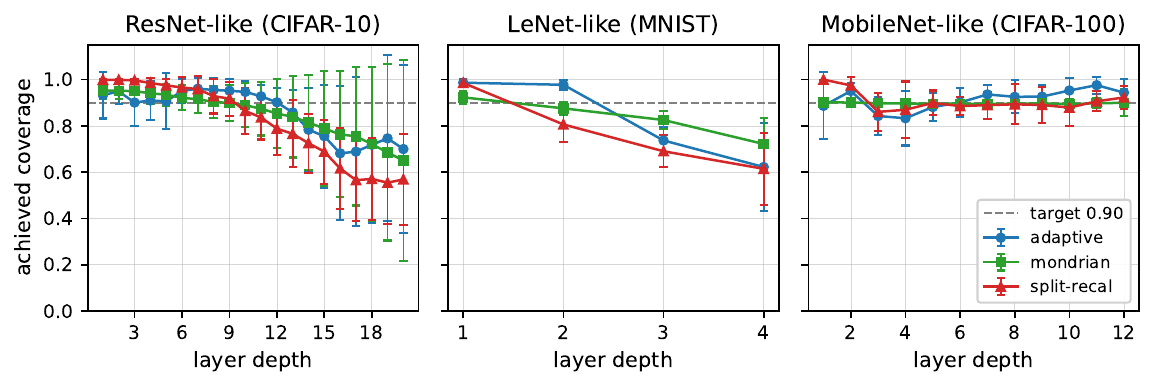}
\caption{
    Coverage by construction depth (target 0.90, no-prune audits, eight seeds): ResNet-like/CIFAR-10 (left), LeNet-like/MNIST (center), MobileNet-like/CIFAR-100 (right).
    Global offsets produce family-dependent coverage imbalances.
    Mondrian ACI reduces these imbalances, with near-target coverage at every depth on MobileNet-like.
}
\label{fig:depth}
\end{figure*}

Sequential construction introduces drift both across episodes, as the policy changes, and within episodes, as partial networks grow.
To identify imbalances hidden by marginal coverage, we audit coverage by construction depth with pruning disabled (Sec.~\ref{sssec:loop}).
Every candidate is evaluated and the filter receives live feedback.
We compare global ACI, per-depth Mondrian ACI, and periodically recalibrated split across all three families at target coverage 0.90.
The audits use eight seeds because each depth receives only a fraction of the observations available for marginal coverage estimates.
Figure~\ref{fig:depth} plots the per-depth coverage curves for the three families, and Table~\ref{tab:deptherr} summarizes the conditional error by depth region.
The target depth range is family-specific (ResNet-like $[8, 20]$, LeNet-like $[2, 4]$, MobileNet-like $[4, 12]$), and coverage is audited at every construction depth up to the terminal.

\begin{table}[t]
\centering
\caption{
    Mean calibration error $|\mathrm{coverage}-0.90|$ by depth region (no-prune audits, eight seeds).
    Mondrian ACI improves conditional coverage, most clearly on LeNet-like and MobileNet-like.
    Sparse deep groups limit the improvement on ResNet-like.
}
\label{tab:deptherr}
\footnotesize
\setlength{\tabcolsep}{4pt}
\begin{tabular}{@{}llccc@{}}
\toprule
family & depth region & adaptive & mondrian & split-recal \\
\midrule
ResNet-like & 1--13 (dense) & 0.034 & \textbf{0.032} & 0.075 \\
 & 14--20 (sparse) & 0.175 & \textbf{0.160} & 0.287 \\
 & all & 0.083 & \textbf{0.077} & 0.149 \\
\midrule
LeNet-like & all (1--4, dense) & 0.152 & \textbf{0.075} & 0.169 \\
MobileNet-like & all (1--12, dense) & 0.039 & \textbf{0.003} & 0.028 \\
\bottomrule
\end{tabular}
\end{table}

On ResNet-like/CIFAR-10, periodically recalibrated split over-covers shallow partial networks (coverage 1.00) and under-covers deeper ones, reaching 0.57 at depth 20.
Global ACI roughly halves the conditional error and stays near target through depth 13.
It still under-covers the deepest layers, with shallow over-coverage compensating in the marginal average.
Mondrian ACI removes shallow over-coverage but remains limited at deeper layers.
Few episodes reach depth 16 or beyond; these sparse groups are initialized from the global offset and then adapt slowly from few observations, with per-depth seed standard deviations reaching 0.4.

Pooling sparse deep groups into a shared depth bin reduces their error from 0.160 to 0.079, supporting the sample-size explanation.
Pooling trades variance for bias because a shared offset cannot track differences in difficulty within a bin.

\textbf{Effect of depth-group sample size.}
LeNet-like/MNIST provides approximately 455 partial networks per seed and at least 54 observations at every depth.
Mondrian ACI is closest to target at every depth, with mean conditional error 0.075, compared with 0.152 for global ACI and 0.169 for periodically recalibrated split.
Global ACI already under-covers at depth 3 in this four-layer space, whose stream is dominated by shallow steps.
Nevertheless, all methods under-cover at the terminal depth (0.61--0.72 coverage with approximately 54 observations per seed), showing that exceeding the calibration minimum does not ensure accurate finite-run coverage.

\textbf{Family-dependent coverage imbalances.}
MobileNet-like/CIFAR-100 is comparatively well sampled across depth, with about 150 networks per seed at depths 1--4 and 17 at depth 12 over the full audit; these totals span the whole run, not only the calibration prefix, so the deeper groups rely mostly on post-warm-up adaptation.
Mondrian ACI achieves coverage 0.893--0.902 across all twelve depths, a range of 0.009, approximately $15\times$ narrower than either global arm.
Its mean conditional error is 0.003, compared with 0.039 for global ACI and 0.028 for periodically recalibrated split.
Here the global arms under-cover at shallow-to-middle depths (0.833--0.870 at depths 3--4) and over-cover deeper partial networks; terminal coverage is on target.
This reverses the depth profile of the other two families, so the difficult depths depend on the family.
Marginal calibration alone consequently gives an incomplete account of filtering behavior during sequential construction.

\begin{figure}[t]
\centering
\includegraphics[width=0.65\columnwidth]{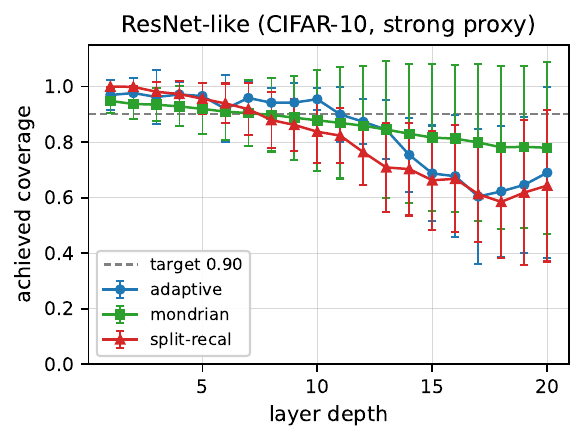}
\caption{
    Coverage by construction depth on the \textit{strong-proxy} ResNet-like/CIFAR-10 setting (target 0.90, no-prune, eight seeds), the setting of the directed-search results.
    The global arms over-cover shallow depths and under-cover at greater depths; per-depth Mondrian tracks the target across depth at roughly half their overall error.
}
\label{fig:depthstrong}
\end{figure}

\textbf{Depth coverage under the stronger proxy.}
We repeat the ResNet-like audit with the strong proxy used for directed search (eight seeds; Fig.~\ref{fig:depthstrong}).
The depth imbalance persists under the stronger proxy, with larger deep-region errors for the global filters: at depths 14--20, coverage is 0.60--0.75 for ACI and 0.59--0.70 for periodically recalibrated split (deep-region errors 0.231 and 0.258).
These runs also visit deep networks more frequently, providing more observations for per-depth adaptation.

Mondrian ACI has the lowest overall error: 0.052, compared with 0.113 for global ACI and 0.140 for periodically recalibrated split.
With more observations in deep groups, its deep-region error falls from 0.160 under the weak proxy to 0.100, even without pooling.
This provides further evidence that group sample size limits adaptation.

\subsection{E5: Controller efficiency relative to uninformed search}
\label{ssec:e5}

The filter is controller-agnostic; this experiment asks what the controller itself contributes.
On ResNet-like with the strong proxy (15 epochs, 30\% of training data), PPO, random search, and regularized evolution each receive 1{,}000 real evaluations, with filtering disabled and three seeds (Table~\ref{tab:budget}).
The stronger proxy avoids the isolated optimum that makes the weak-proxy comparison depend mainly on warm-up sampling.

\begin{table}[t]
\centering
\caption{
    Best reward vs evaluation budget on the standard landscape (512\,kB budget, no filtering; mean over seeds).
    This is a different, less-constrained testbed than the 256\,kB one of Table~\ref{tab:acq}, so absolute rewards are not comparable across the two.
}
\label{tab:budget}
\footnotesize
\setlength{\tabcolsep}{3pt}
\begin{tabular}{@{}lccccc@{}}
\toprule
evaluations & 100 & 200 & 500 & 1000 & final ($\pm$ std) \\
\midrule
random & 0.699 & 0.719 & 0.725 & \textbf{0.725} & $\mathbf{0.7252 \pm 0.0035}$ \\
evolution & 0.707 & 0.713 & 0.715 & 0.715 & $0.7148 \pm 0.0117$ \\
ppo & 0.704 & 0.704 & 0.704 & 0.704 & $0.7040 \pm 0.0058$ \\
\bottomrule
\end{tabular}
\end{table}

PPO has no early-budget advantage.
Its best reward stops improving around evaluation 100, both baselines overtake it by evaluation 200, and its final reward trails random search by 0.021 and regularized evolution by 0.011.
Its mean proposal reward nevertheless rises from 0.25 to 0.40 while random search's remains flat.
Policy concentration improves average proposals but can reduce exploration of rare high-reward candidates.
The result agrees with previous random-search comparisons \citep{Li2020d} and motivates testing whether optimistic acquisition improves best-candidate search.

\subsection{E6: Directed search with calibrated acquisition}
\label{ssec:e6}

As a preliminary study, we also use the calibrated bound as an acquisition function on a single constrained testbed (the ResNet-like per-layer space, strong proxy, 256\,kB budget, three seeds).
The tighter 256\,kB budget (versus the 512\,kB of the controller comparison in Table~\ref{tab:budget}) makes feasibility binding, producing a landscape on which the search methods are distinguishable; absolute rewards are therefore not comparable across the two testbeds.
Greedy acquisition (Eq.~\eqref{eq:greedy}) outperforms random search at each reported budget through 1{,}000 evaluations, with gains of approximately 0.040 in reward and 0.061 in accuracy at 1{,}000.
Decomposing the $0.039$ gain over random search (0.656): surrogate-mean selection alone reaches 0.682 (a $0.026$ gain, the largest share), fixed optimism adds $0.009$ (0.691), and online calibration a further $0.004$ (0.695).
Most of the improvement therefore persists without online calibration: surrogate-based selection provides the largest observed gain, with additional gains from optimism.
Across three seeds, reward standard deviations were approximately $0.002$ for both optimistic arms and $0.010$ for the surrogate-mean arm; these estimates are insufficient to establish a reliable variance reduction.

MCTS outperforms random search at 100--200 evaluations but plateaus by 500 and finishes at or below random.
MCTS performance changes little across $c_{\text{puct}} \in \{0.5, 1.4, 3.0\}$, with best reward varying by approximately 0.01 at each evaluation budget.

Reward shaping (Eq.~\eqref{eq:shaped}) does not reliably beat the unshaped controller (median reward $0.67$ vs $0.63$, accuracy $0.70$ vs $0.72$, within seed noise) and can destabilize training: in a separate eight-seed experiment, two seeds diverge into infeasible regions.
These results are preliminary because they cover one family and testbed; Table~\ref{tab:acq} reports the full comparison.

\begin{table}[t]
\centering
\caption{
    Directed search on the constrained testbed (ResNet-like, strong proxy, 256\,kB; three-seed means): best reward at each budget and best accuracy at 1{,}000, with across-seed standard deviations as subscripts.
    Acquisition, MCTS and random commit only feasible candidates; the terminal-only PPO controller explores infeasible candidates under a soft penalty, so only its final best-\textit{feasible} reward is comparable ($-$ marks omitted intermediate budgets).
}
\label{tab:acq}
\footnotesize
\setlength{\tabcolsep}{3pt}
\begin{tabular}{@{}lccccc@{}}
\toprule
arm & 100 & 200 & 500 & 1000 & acc.\ @1000 \\
\midrule
acq.\ [calibrated] & 0.622 & 0.661 & 0.689 & \textbf{0.695}$_{\pm.002}$ & \textbf{0.734}$_{\pm.011}$ \\
acq.\ [uncalibrated] & 0.629 & 0.654 & 0.676 & 0.691$_{\pm.002}$ & 0.727$_{\pm.014}$ \\
acq.\ [mean] & 0.629 & 0.657 & 0.682 & 0.682$_{\pm.010}$ & 0.723$_{\pm.017}$ \\
mcts [calibrated] & 0.604 & 0.627 & 0.633 & 0.637$_{\pm.004}$ & 0.661$_{\pm.016}$ \\
mcts [uncalibrated] & 0.631 & 0.634 & 0.642 & 0.653$_{\pm.020}$ & 0.669$_{\pm.023}$ \\
ppo (terminal) & $-$ & $-$ & $-$ & 0.659$_{\pm.039}$ & 0.686$_{\pm.042}$ \\
random & 0.520 & 0.604 & 0.620 & 0.656$_{\pm.026}$ & 0.673$_{\pm.034}$ \\
\bottomrule
\end{tabular}
\end{table}

\section{Conclusion}
\label{sec:conclusion}
This study characterizes how adaptive coverage control relates to pruning efficiency and search quality in hardware-aware NAS.
A common experimental framework supports comparisons across uncertainty models, proposal mechanisms, three CNN families, and a tabular benchmark with measured hardware costs.
ACI provides steerable update-time coverage on evaluated outcomes where static calibration deviates from its target.
Across three CNN families, it prunes 25--50\% of evaluations without measured loss in partial-training proxy accuracy at the chosen operating thresholds.
AgACI achieves similar tracking without selecting a single step size.

Sequential audits reveal depth-dependent imbalances that marginal coverage conceals.
Mondrian adaptation reduces mean conditional error from 0.152 to 0.075 on LeNet-like and from 0.039 to 0.003 on MobileNet-like networks, while improvement on ResNet-like is limited by sparse deep groups.
These results establish the value of evaluating calibration at the construction depths where decisions are made.

The cross-fidelity benchmark quantifies how pruning thresholds affect full-fidelity false discards.
Matching thresholds to the proxy scale reduces these losses at the cost of conservative pruning; comparisons at matched pruning fractions expose differences from raw-proxy ranking.
The tested filters, including GP-UCB, exhibit broadly similar trade-offs.

The RL controller does not outperform random search at matched budgets despite improving its mean proposal reward.
On one constrained testbed, greedy acquisition improves on random search, with surrogate-based selection providing the largest observed gain and smaller gains from optimism and online calibration.

\subsection{Limitations}

The analytical CNN hardware model supports conclusions about relative costs and budget constraints, not absolute latencies.
The tabular benchmark uses measured HW-NAS-Bench costs, but direct deployment validation remains future work.
Coverage on partial-training outcomes does not establish calibration to fully trained rewards, and solution quality depends on the proxy's ranking fidelity.
The ResNet-like controller and acquisition comparisons require the stronger proxy, and acceptance thresholds are selected separately for each family.
Mondrian adaptation initializes sparse groups from the global offset before they adapt from their own observations; even groups above the calibration minimum can show finite-run miscoverage.
Live filtered runs reveal coverage only on accepted candidates.
No-prune audits and tabular reward lookups permit uncensored evaluation, but correcting selection bias in live training remains open.
The experiments cover from-scratch CNN classification and a tabular NAS benchmark; transfer to transformers, fine-tuning, detection, and segmentation remains untested.
Three-seed calibration sweeps and eight-seed depth audits establish the main empirical effects but leave substantial uncertainty at sparse depths and small evaluation budgets.
The acquisition study is limited to one family and testbed and is separate from the main filtering contribution.

\subsection{Outlook}

Priorities are to validate optimistic acquisition across broader search spaces, improve conditional coverage in sparse groups, and evaluate calibration against fully trained rewards and physical-device measurements.
The same bounds may also inform evaluation-resource allocation and stopping rules.
A further priority is direct control of discard risk under censored feedback and search-induced distribution shift, using conformal selection or risk-controlling calibration.

\appendix
\counterwithin{figure}{section}
\counterwithin{table}{section}
\counterwithin{equation}{section}
\counterwithin{algorithm}{section}
\section{Implementation settings}
\label{app:settings}

Table~\ref{tab:settings} collects the default settings used across all experiments.
The three CNN families share the warm-up budget (100 candidates in filtered search) and, for the MLP-based arms, a three-layer multilayer perceptron surrogate trained by mean-squared error on the candidate features, with split-conformal calibration (half the available data fits the surrogate, the other half forms the held-out nonconformity scores).
The adaptive, periodically recalibrated split, and calibrate-once split filters differ in their use of post-warm-up feedback: the adaptive filter updates its conformal offset after every evaluation, the periodically-recalibrated filter re-fits every $P = 50$ accepted candidates, and the calibrate-once filter is frozen after warm-up.
The acceptance threshold $\tau$ is calibrated per family to the reward distribution of an unfiltered probe run (a location within that distribution, not a transferable constant).
The stronger proxy $(E,\rho) = (15, 0.3)$ replaces the default $(5, 0.1)$ for the MobileNet-like calibration and replay experiments, the ResNet-like controller and acquisition comparisons, and the additional ResNet-like depth audit.
The NAS-Bench-201 tabular benchmark has its own settings (final table row): the observed 12-epoch and reference 200-epoch CIFAR-100 rewards, measured HW-NAS-Bench \texttt{edgegpu} latency and energy each normalized by its table maximum, and the pruning-threshold ($\tau_{\text{proxy}}$) and $\delta$ sweep grids.

\begin{table*}[t]
\centering
\caption{Default implementation settings.}
\label{tab:settings}
\footnotesize
\begin{tabular}{@{}p{0.20\textwidth}p{0.60\textwidth}@{}}
\toprule
Component & Setting \\
\midrule
Reward & $\lambda = 0.2$, $\nu = 1.0$ \\
Proxy & $(E,\rho) = (5, 0.1)$; $(15, 0.3)$ for MobileNet-like calibration/replay, ResNet-like controller/acquisition comparisons, and the additional ResNet-like depth audit \\
Surrogate & MLP $d \to 64 \to 64 \to 1$, ReLU; MSE, Adam, lr $10^{-2}$, 300 epochs \\
Calibration split & $1/2$ surrogate fit, $1/2$ score quantile \\
Warm-up & 100 candidates (filtered search); $W = 40$ (acquisition search) \\
ACI & $\gamma = 0.05$, constant \\
AgACI & grid $\{0.01, 0.02, 0.05, 0.1, 0.2\}$; anytime Hedge \\
Recalibrated split & period $P = 50$ \\
Mondrian & group = layer depth; $n_{\min} = 10$, global fallback \\
Normalized & MAD scale, floor $\varepsilon_u = 10^{-3}$ \\
GP baseline & RBF + white kernel, online re-fit, $\kappa = \Phi^{-1}(1-\delta)$ \\
PPO & $\epsilon = 0.2$, $c_v = 0.5$, $c_H = 0.01$, 4 epochs, lr $5 \times 10^{-4}$; $\gamma_{\text{RL}} = 0.99$ \\
Rejection reward & $r_{\text{rej}} = -1$ \\
Acquisition & pool $S = 64$; $N_{\text{sim}} = 200$, $c_{\text{puct}} = 1.4$; $\beta = 0.1$ \\
NAS-Bench-201 & Observed reward: NATS-Bench 12-epoch CIFAR-100 accuracy; reference reward: 200-epoch accuracy.
HW-NAS-Bench \texttt{edgegpu} latency+energy, each normalized by its table maximum, $\lambda_{\text{lat}} = \lambda_{\text{en}} = 0.2$.
Good threshold $\tau_{\text{ref}} = 0.438$ (200-epoch median); pruning threshold $\tau_{\text{proxy}} \in \{.214, .25, .30, .35, .40, .438\}$.
$1{,}000$ cells/run ($50 \times 20$), $100$-cell warm-up, false discards scored post-warm-up; 20 seeds, $\delta \in \{.05, .1, .2, .3, .4\}$.
Synthetic-noise diagnostic (Appendix~\ref{app:noise}) instead observes the 200-epoch reward plus noise $\sigma \in \{0, .05, .1\}$. \\
\bottomrule
\end{tabular}
\end{table*}

\section{Telescoping coverage bound for ACI}
\label{app:proof}

This appendix gives a self-contained derivation of the long-run coverage guarantee behind Eq.~\eqref{eq:telescope}, as a specialization of the standard Adaptive Conformal Inference argument of Gibbs and Cand\`es; it also makes the constant $C$ of Eq.~\eqref{eq:telescope} explicit.

\textbf{Setup.}
At step $t$ an evaluated candidate $x_t$ is scored by the surrogate $g(x_t)$, and its bound $U_t = g(x_t) + \alpha_t$ is formed \emph{before} the real evaluation returns a reward $r_t$.
The signed residual and miscoverage indicator are
\begin{equation}
\begin{aligned}
\xi_t &:= r_t - g(x_t),\\
\err_t &:= \ind\{\xi_t > \alpha_t\} = \ind\{r_t > U_t\} \in \{0,1\}.
\end{aligned}
\end{equation}
The index $t$ ranges over the evaluated candidates that receive an update, and the offset recursion (Eq.~\eqref{eq:aci}), with constant step size $\gamma>0$ and target level $\delta\in(0,1)$,
\begin{equation}
\label{eq:sm-aci}
\alpha_{t+1} = \alpha_t + \gamma\,(\err_t - \delta)
\end{equation}
is applied \emph{without interruption}: no offset resets and no projection between steps.
It raises $\alpha$ by $\gamma(1-\delta)$ on a miscoverage and lowers it by $\gamma\delta$ otherwise, so increases and decreases balance when the average miscoverage equals $\delta$; the individual offsets need not converge.

\textbf{Bounded residuals.} Assume finite $\xi_{\min}\le\xi_{\max}$ with $\xi_t\in[\xi_{\min},\xi_{\max}]$ for all $t$.
In the fixed-surrogate ACI configurations (the \texttt{adaptive} and \texttt{normalized} filters, whose surrogate is fit once at warm-up and then held fixed) this holds structurally: the design space $\mathcal{A}$ is finite and $g$ is evaluated only on $\mathcal{A}$, so its range is finite, and the reward is bounded, $r_t\in[R_{\min},R_{\max}]$, over the finite $\mathcal{A}$; hence $\xi_t$ is bounded with no further assumption.
For any configuration that re-fits the surrogate over time, the same guarantee requires a \emph{uniform} residual bound across refits, $\sup_t|\xi_t|<\infty$.
If a surrogate's range were not otherwise controlled, clipping the prediction to the reward interval, $\tilde g=\operatorname{clip}(g,R_{\min},R_{\max})$, is an optional device that enforces bounded residuals: since $r_t\in[R_{\min},R_{\max}]$, one has $|r_t-\tilde g(x_t)|\le|r_t-g(x_t)|$, so clipping never enlarges the residual (it may change $U_t$, but not loosen the bound on $|\xi_t|$).

\begin{proposition}[Telescoping coverage bound]
\label{prop:sm-telescope}
For every horizon $T\ge1$, the empirical miscoverage rate satisfies the exact identity
\begin{equation}
\label{eq:sm-ident}
\Big|\tfrac{1}{T}\textstyle\sum_{t=1}^{T}\err_t-\delta\Big| = \frac{|\alpha_{T+1}-\alpha_1|}{\gamma T}.
\end{equation}
For any finite initialization $\alpha_1$, the offset stays in $[L,H]$ with
\begin{equation}
\label{eq:sm-interval}
\begin{aligned}
L&=\min\{\alpha_1,\ \xi_{\min}-\gamma\delta\},\\
H&=\max\{\alpha_1,\ \xi_{\max}+\gamma(1-\delta)\},
\end{aligned}
\end{equation}
so, writing $C:=H-L$,
\begin{equation}
\label{eq:sm-bound}
\Big|\tfrac{1}{T}\textstyle\sum_{t=1}^{T}\err_t-\delta\Big| \le \frac{C}{\gamma T}\xrightarrow[T\to\infty]{}0 .
\end{equation}
Long-run empirical coverage $\tfrac1T\sum_t(1-\err_t)$ therefore converges to $1-\delta$ for an arbitrary sequence $\{(x_t,r_t)\}$, with no exchangeability, stationarity, or distributional assumption.
\end{proposition}

\begin{proof}
Summing Eq.~\eqref{eq:sm-aci} over $t=1,\dots,T$ telescopes the left side: $\alpha_{T+1}-\alpha_1=\gamma\sum_t(\err_t-\delta)$.
Dividing by $\gamma T$ and taking absolute values gives Eq.~\eqref{eq:sm-ident}.
For the interval, the update is self-correcting.
The offset rises only when $\err_t=1$, i.e.\ $\xi_t>\alpha_t$; once $\alpha_t\ge\xi_{\max}$ one has $\xi_t\le\xi_{\max}\le\alpha_t$, forcing $\err_t=0$ and a down-step, so $\alpha$ exceeds $\xi_{\max}$ by at most a single up-step, $\alpha_t\le\xi_{\max}+\gamma(1-\delta)$, unless it started higher.
Symmetrically $\alpha_t\ge\xi_{\min}-\gamma\delta$ unless it started lower.
Hence $\alpha_t\in[L,H]$ and $|\alpha_{T+1}-\alpha_1|\le H-L=C$; substituting into Eq.~\eqref{eq:sm-ident} yields Eq.~\eqref{eq:sm-bound}.
\end{proof}

\textbf{What the bound covers.}
The index $t$ ranges over evaluated candidates whose bound is formed before their reward is seen, and the recursion is uninterrupted.
The statement therefore controls the cumulative coverage of the \emph{observed} (evaluated) stream only.
It says nothing about pruned candidates, whose rewards are not used for filter updates, and does not by itself imply a conditional false-discard guarantee.
Coverage measurements vary along three independent axes.
\emph{Population}: the \emph{evaluated} stream (candidates actually evaluated) or the \emph{full proposal} stream (all proposed candidates, including pruned ones, recoverable by benchmark lookup or complete replay).
\emph{Timing}: \emph{update-time} bounds, formed with the offset updated between accepted evaluations, or \emph{decision-time} bounds, formed with the offset held fixed at the start of a batch and used to prune.
\emph{Conditioning}: marginal coverage, or \emph{depth-conditional} coverage grouped by construction depth (the Mondrian arm).
The bandit runs accept a batch of $B = 20$ proposed candidates at the pre-batch offset, then update the offset on the accepted candidates in sequence, so update-time and decision-time coverage can differ within a batch and are guaranteed to coincide at $B = 1$.
We report update-time coverage on the evaluated stream, marginal or depth-conditional as noted.

\begin{corollary}[Normalized ACI]
With the normalized score $\xi_t=(r_t-g_t(x_t))/u_t(x_t)$ and the multiplier update of the main text in place of Eq.~\eqref{eq:sm-aci}, Proposition~\ref{prop:sm-telescope} applies verbatim provided the normalized residuals are bounded.
The positive scale floor $u_t\ge\varepsilon_u>0$ supplies this whenever the raw residuals are bounded: if $|r_t-g_t(x_t)|\le B<\infty$ and $u_t\ge\varepsilon_u$, then $|\xi_t|\le B/\varepsilon_u$ (with $B=R_{\max}-R_{\min}$ in the special case of predictions clipped to the reward interval).
\end{corollary}

\begin{corollary}[Mondrian ACI]
Applying Proposition~\ref{prop:sm-telescope} separately to each group $k$, with its own offset and $T$ replaced by the group's observed count $T_k$, gives, for every $T_k\ge1$, $\big|\tfrac{1}{T_k}\sum\err-\delta\big|\le C_k/(\gamma T_k)$, where $C_k=H_k-L_k$ is the group's own interval width (a shared $C\ge\max_k C_k$ is also valid).
Per-group convergence requires $T_k\to\infty$, so sparsely sampled groups carry weak finite-run bounds; a global initialization is admissible because Proposition~\ref{prop:sm-telescope} allows any finite $\alpha_1$.
\end{corollary}

\section{Consolidated algorithms and calibration-state management}
\label{app:algos}

Algorithm~\ref{alg:filter} is the shared filtered-search driver; the conformal arms listed in Table~\ref{tab:policies} differ only in their \textsc{Calibrate}/\textsc{Update} policies (the GP-UCB baseline and the no-filter arm are not conformal and are omitted).
Algorithm~\ref{alg:acq} is the greedy acquisition procedure; the MCTS and optimism-shaped-controller variants share its calibration-state update (calibrate at warm-up, then one online \textsc{Update} per real evaluation on a fixed surrogate), while their proposal generation, controller training, and evaluation granularity differ; the shaped controller also uses a different training reward and partial-network evaluations.
We write $\textsc{UpperBound}_F(x) = g(x) + \alpha$ for the bound (normalized $g + q\,u$; Mondrian $g + \alpha_{k(x)}$), $\err_t = \ind\{r_t > \textsc{UpperBound}(x_t)\}$ for the one-sided miscoverage indicator, and $\alpha$ for the offset ($q$ for the normalized arm).

\begin{algorithm}[t]
\caption{Conformal-filtered search (shared driver; controller policy updates omitted).}
\label{alg:filter}
\begin{algorithmic}[1]
\Require warm-up $(X_w, r_w)$, threshold $\tau$, filter $F$, batch size $B$
\State $\textsc{Calibrate}_F(X_w, r_w)$
\For{each post-warm-up batch $\{x_1, \dots, x_B\}$}
  \State $a \gets \textsc{Accepts}_F(\{x_i\}, \tau)$ \Comment{whole batch decided from the current bound}
  \For{each $i$ with $a_i$ true, in order}
    \State evaluate $r_i$; \quad $\textsc{Update}_F(x_i, r_i)$ \Comment{Table~\ref{tab:policies}}
  \EndFor
  \For{each $i$ with $a_i$ false}
    \State prune $x_i$; controller receives $r_{\text{rej}}$ \Comment{no reward observation or filter update}
  \EndFor
\EndFor
\end{algorithmic}
\end{algorithm}

\begin{table*}[!t]
\centering
\caption{
    Per-arm \textsc{Update} policies and surrogate/offset handling in Algorithm~\ref{alg:filter} (surrogates remain fixed after warm-up except for split-recal, which refits; adaptive offsets continue updating, while split-recal periodically resets its offset), and whether the online offset recursion is left uninterrupted (so the telescoping guarantee of Proposition~\ref{prop:sm-telescope} applies) or reset.
    For AgACI the guarantee applies to each expert, not to the aggregated offset.
}
\label{tab:policies}
\footnotesize
\begin{tabular}{@{}l p{0.5\textwidth} l@{}}
\toprule
arm & \textsc{Update}$(x_t, r_t)$ & surrogate / offset \\
\midrule
static (split) & no-op & fixed / fixed \\
split-recal & buffer; every $P$, recompute $g$ and the offset quantile & refit $P$ / \textbf{reset} $P$ \\
adaptive (ACI) & $\alpha \mathrel{+}= \gamma(\err_t - \delta)$ & fixed / uninterrupted \\
agaci & each expert ACI-steps on its own pre-update bound; Hedge-reweight by per-step pinball loss (pre-update bounds); $\alpha$ is their weighted mean & fixed / uninterrupted per expert \\
normalized & $q \mathrel{+}= \gamma(\err_t - \delta)$, bound $g + q\,u$ & fixed / uninterrupted \\
mondrian & $\alpha_{k(x_t)} \mathrel{+}= \gamma(\err_t - \delta)$; global $\alpha$ for unseen groups & fixed / uninterrupted per group \\
\bottomrule
\end{tabular}
\end{table*}

\begin{algorithm}[t]
\caption{Greedy calibrated acquisition (reported protocol; fixed surrogate).}
\label{alg:acq}
\begin{algorithmic}[1]
\Require warm-up budget $W$, total budget $N$, pool size $S$, filter $F$
\State evaluate $W$ random candidates; \quad $\textsc{Calibrate}_F$ on them \Comment{surrogate fixed hereafter}
\For{$\text{step} = 1, \dots, N - W$}
  \State sample a pool $\{x^{(1)}, \dots, x^{(S)}\}$ and keep its feasible subset (within budget $M_B$)
  \State $i^\star \gets \arg\max_{i:\, x^{(i)}\ \text{feasible}} \textsc{UpperBound}_F(x^{(i)})$; \quad $x \gets x^{(i^\star)}$
  \State evaluate $r$; \quad $\textsc{Update}_F(x, r)$ \Comment{online ACI step}
\EndFor
\State \Return best evaluated candidate
\end{algorithmic}
\end{algorithm}

\textbf{Calibration-state management.}
For \texttt{adaptive}, \texttt{normalized}, \texttt{mondrian} and the reported greedy acquisition, the surrogate is fit once at warm-up and held fixed, and the offset recursion is \emph{uninterrupted}, so Proposition~\ref{prop:sm-telescope} applies (per group for \texttt{mondrian}; \texttt{agaci} satisfies it per expert, but its aggregated offset carries no telescoping guarantee).
The offset is \emph{reset}, that is, reinitialized from a fresh residual quantile, only for \texttt{split-recal} (every $P$); the GP-UCB baseline instead re-fits its model online.
Every online \textsc{Update} forms its miscoverage indicator from the current surrogate and offset, $\err_t = \ind\{r_t > g(x_t) + \alpha\}$ (normalized: $g + q\,u$; AgACI: each expert against its own pre-update bound).
Calibration uses \emph{absolute} residuals $|r - g(x)|$ for the one-sided upper bound.
The clipping device of the bounded-residual discussion is \emph{not} used in our runs: in the fixed-surrogate filters the residuals are bounded structurally (finite design space, bounded reward).
The normalized multiplier $q$ is not constrained to be nonnegative; a negative $q$ would turn the bonus $q\,u$ into an uncertainty penalty, and we do not clip it.

\section{Pruning-threshold trade-off under the 12-epoch proxy}
\label{app:gap}

\begin{figure*}[t]
\centering
\includegraphics[width=\textwidth]{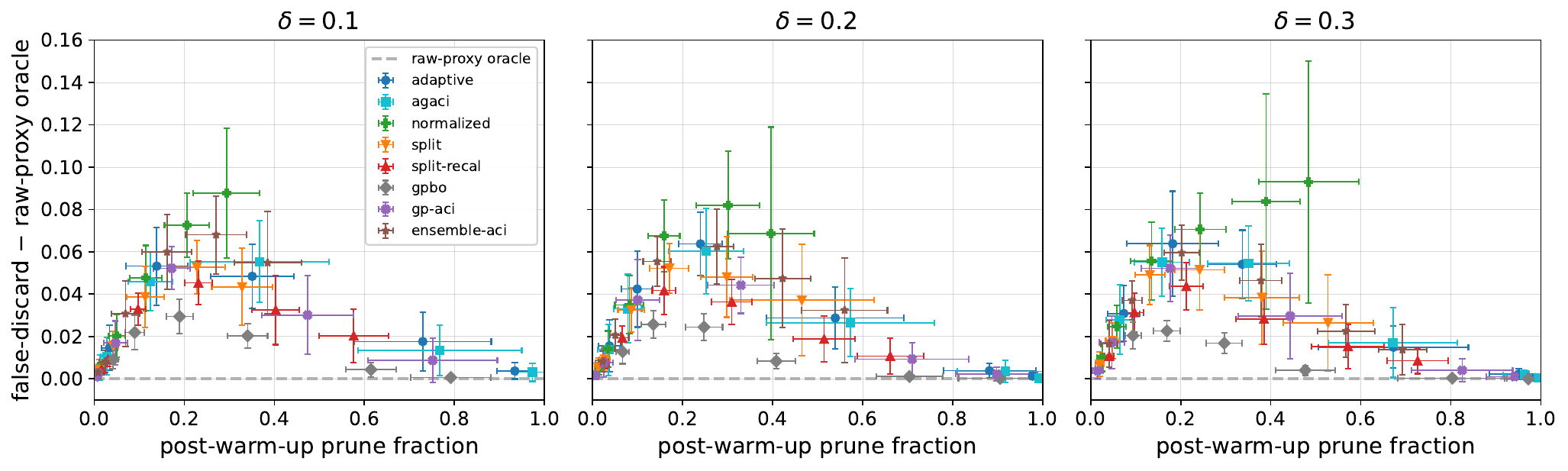}
\caption{
    Paired gaps to the retrospective raw-proxy ranking reference for the operating points in the E3 trade-off figure (Fig.~\ref{fig:nb201tradeoff}): each seed's filter false-discard rate minus the reference rate at the same pruning fraction.
    Markers show seed means; bars show across-seed standard deviations of pruning fractions and paired gaps.
    The horizontal zero line denotes equal false-discard rates.
    Across all $240$ operating points, the mean excess false-discard rate relative to this reference is $2.6$ percentage points (Table~\ref{tab:nb201gap}).
}
\label{fig:nb201gap}
\end{figure*}

Figure~\ref{fig:nb201gap} complements the E3 trade-off figure (Fig.~\ref{fig:nb201tradeoff}) by comparing the filters with raw-proxy ranking under the 12-epoch proxy.
We sweep the pruning threshold $\tau_{\mathrm{proxy}}$ from the 12-epoch reward median ($0.214$) to the full-fidelity median ($0.438$), keeping the good-candidate threshold fixed at $\tau_{\mathrm{ref}} = 0.438$.
Both pruning fractions and false-discard rates exclude warm-up candidates.

The reference is a \emph{retrospective raw-proxy ranking} of each run's own post-warm-up candidates.
It discards the same number of cells as the filter, selecting those with the lowest tabulated 12-epoch rewards, and scores false discards against the 200-epoch rewards.
This requires proxy outcomes for every candidate, including those the filter never evaluated.
The reference is not a bound on achievable full-fidelity false-discard rates: a feature-based predictor could rank candidates better than the raw proxy.
The line $\mathrm{FD} = \delta$ compares the observed false-discard rate with the nominal miscoverage parameter; it is a numerical reference, not a guaranteed false-discard bound.
Table~\ref{tab:nb201gap} summarizes paired gaps to this reference, computed per seed at matched pruning fractions and then averaged for each operating point.
No filter consistently beats the reference; the operating-point mean gaps average $2.6$ percentage points and reach at most $9.3$ points.

\begin{table}[t]
\centering
\caption{
    Paired gaps to the retrospective raw-proxy ranking reference, summarized per filter over $n$ operating points (pruning threshold $\times$ $\delta$).
    Each operating-point gap is averaged over seeds; the table reports the mean and maximum of these averages, not a bound on individual runs.
    Positive values indicate more good candidates discarded by the filter at the same pruning fraction.
}
\label{tab:nb201gap}
\footnotesize
\begin{tabular}{@{}lrrr@{}}
\toprule
filter & $n$ & mean gap & max gap \\
\midrule
adaptive      & 30 & $+0.025$ & $+0.064$ \\
agaci         & 30 & $+0.024$ & $+0.067$ \\
normalized    & 30 & $+0.044$ & $+0.093$ \\
split         & 30 & $+0.029$ & $+0.056$ \\
split-recal   & 30 & $+0.023$ & $+0.045$ \\
gpbo          & 30 & $+0.013$ & $+0.035$ \\
gp-aci        & 30 & $+0.019$ & $+0.057$ \\
ensemble-aci  & 30 & $+0.035$ & $+0.073$ \\
\midrule
all           & 240 & $+0.026$ & $+0.093$ \\
\bottomrule
\end{tabular}
\end{table}

\section{Tabular benchmark under synthetic proxy noise}
\label{app:noise}

\begin{figure*}[t]
\centering
\includegraphics[width=\textwidth]{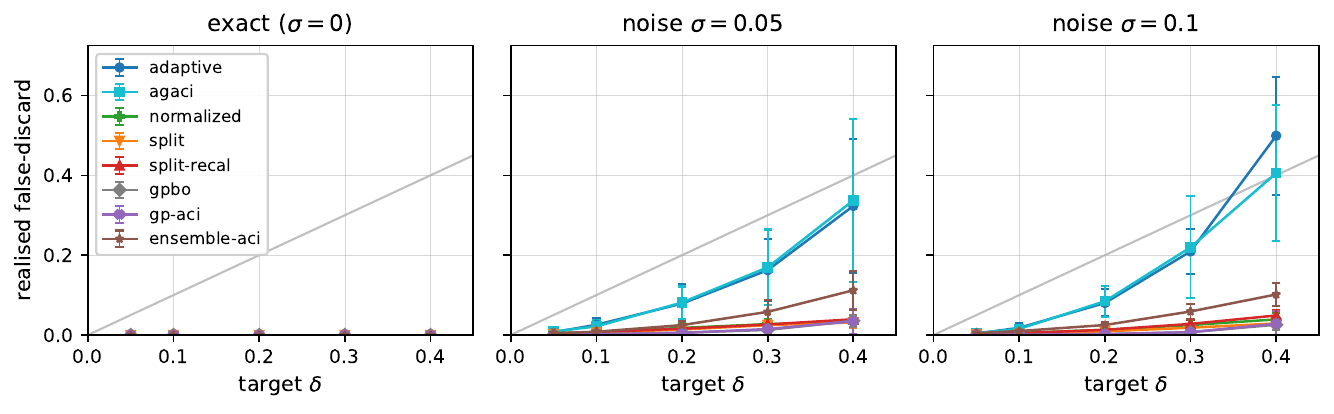}
\caption{
    Fraction of truly good candidates pruned vs requested miscoverage $\delta$ on NAS-Bench-201/CIFAR-100 with measured HW-NAS-Bench edge-GPU costs (20 seeds; points = seed means, bars = standard deviations).
    The grey line marks equality with $\delta$, an empirical comparison rather than a conditional-coverage guarantee.
    Left: exact proxy, every arm at zero.
    Center: at $\sigma=0.05$, all arms remain below the reference line.
    Right: at $\sigma=0.1$, ACI and AgACI exceed the reference at loose $\delta$, while normalized ACI and its GP- and ensemble-backed variants remain below it.
}
\label{fig:nb201}
\end{figure*}

This diagnostic varies proxy reliability through synthetic noise, complementing the realistic 12-epoch proxy of Sec.~\ref{ssec:e3}.
On NAS-Bench-201/CIFAR-100 with measured HW-NAS-Bench edge-GPU costs ($\tau = 0.438$, $20$ seeds), the controller and filter observe either the exact reward or an estimate with added Gaussian noise, $\sigma \in \{0.05, 0.1\}$.
False discards are always scored against the true tabulated reward.
Alongside normalized ACI, we evaluate GP-backed ACI (\texttt{gp-aci}) and ensemble-backed ACI (\texttt{ensemble-aci}).
Figure~\ref{fig:nb201} plots the realized false-discard rate against the requested $\delta$ in each regime.

\textbf{No false discards with exact rewards.}
With exact rewards, every arm has zero false discards at every tested $\delta$ (Figure~\ref{fig:nb201}, left).
Up to 38\% of proposals are pruned without discarding a good cell, and all arms reach reward 0.609, close to the global optimum of 0.611.
This directly checks the rewards of pruned candidates, complementing the live solution-quality comparison of Sec.~\ref{ssec:e2}.

\textbf{False discards under moderate proxy noise.}
With noisy feedback, the filter can underestimate a good candidate and prune it.
At $\sigma=0.05$, false-discard rates rise with $\delta$ but remain below it for every arm (Figure~\ref{fig:nb201}, center).
This is an empirical result on true rewards, distinct from coverage calibration on the observed proxy.
ACI and AgACI respond most closely to the target, reaching a false-discard rate of approximately 0.32 at $\delta=0.4$.

\textbf{Candidate-dependent widths under heavier noise.}
At $\sigma=0.1$ and $\delta=0.4$, ACI and AgACI prune 0.50 and 0.41 of good cells, respectively, while discarding over half of all proposals (Figure~\ref{fig:nb201}, right).
Normalized ACI, GP-backed ACI, and ensemble-backed ACI remain below $\delta$ throughout.
Calibrate-once and periodically recalibrated split also remain below it, despite using global offsets.
Thus, candidate-dependent widths improve robustness over global ACI in this experiment, but are not the only way to obtain low false-discard rates.
Ensemble-backed ACI offers a useful balance: at both noise levels its false-discard rate remains below $\delta$, and at $\delta=0.4$, $\sigma=0.1$ it prunes 0.21 of below-threshold cells.

\textbf{Comparison with fixed GP-UCB.}
Here ``fixed'' refers to the acquisition multiplier $\kappa$, which is held constant; the GP itself is refit online after each accepted candidate.
This GP-UCB baseline also performs well without conformal calibration.
At $\delta=0.4$, $\sigma=0.1$, it prunes 0.17 of below-threshold cells with a false-discard rate of 0.026, matching or outperforming GP-backed ACI on these measures.
The conformal filters therefore do not uniformly outperform fixed GP-UCB in pruning efficiency.
Their main contribution is online control of evaluated-stream coverage (main text); this diagnostic separately shows the empirical robustness of locally adaptive variants to proxy noise.

\vskip 0.2in
\bibliography{biblio}

\end{document}